\pdfoutput=1
\documentclass[11pt]{article}
\usepackage[preprint]{coling}
\usepackage{times}
\usepackage{latexsym}
\usepackage[T1]{fontenc}
\usepackage[utf8]{inputenc}
\usepackage{microtype}
\usepackage{inconsolata}
\usepackage{graphicx}
\usepackage{amsmath}
\usepackage{amssymb}
\usepackage{amsfonts}
\usepackage{amsthm}
\usepackage{multirow}
\usepackage{array}
\usepackage{booktabs}
\usepackage{subcaption}
\usepackage{adjustbox}
\usepackage{bm}
 
\usepackage{stmaryrd} 

\usepackage{algorithm,algpseudocode}

\newcommand\blfootnote[1]{
  \begingroup
  \renewcommand\thefootnote{}\footnote{#1}
  \addtocounter{footnote}{-1}
  \endgroup
}

\newcommand{\relmiddle}[1]{\mathrel{}\middle#1\mathrel{}}

\usepackage{comment}

\title{Norm of Mean Contextualized Embeddings Determines their Variance}

\author{
  Hiroaki Yamagiwa$^{1}$ \qquad 
  Hidetoshi Shimodaira$^{1,2}$ \\
  ${}^1$Kyoto University \qquad ${}^2$RIKEN\\
  \texttt{hiroaki.yamagiwa@sys.i.kyoto-u.ac.jp,shimo@i.kyoto-u.ac.jp}\\
}

\begin{document}
\maketitle
\begin{abstract}
Contextualized embeddings vary by context, even for the same token, and form a distribution in the embedding space. To analyze this distribution, we focus on the norm of the mean embedding and the variance of the embeddings. In this study, we first demonstrate that these values follow the well-known formula for variance in statistics and provide an efficient sequential computation method. Then, by observing embeddings from intermediate layers of several Transformer models, we found a strong trade-off relationship between the norm and the variance: as the mean embedding becomes closer to the origin, the variance increases. This trade-off is likely influenced by the layer normalization mechanism used in Transformer models. Furthermore, when the sets of token embeddings are treated as clusters, we show that the variance of the entire embedding set can theoretically be decomposed into the within-cluster variance and the between-cluster variance. We found experimentally that as the layers of Transformer models deepen, the embeddings move farther from the origin, the between-cluster variance relatively decreases, and the within-cluster variance relatively increases. These results are consistent with existing studies on the anisotropy of the embedding spaces across layers.
\blfootnote{
Our code is available at \url{https://github.com/ymgw55/Norm-and-Variance}.
}
\end{abstract}

\section{Introduction}\label{sec:intro}
Contextualized embedding is a method for dynamically computing the embeddings of tokens in a sentence. 
Unlike static embeddings such as Skip-gram~\cite{DBLP:conf/nips/MikolovSCCD13} and GloVe~\cite{DBLP:conf/emnlp/PenningtonSM14}, where a predefined embedding is assigned to each word, models such as BERT~\cite{DBLP:conf/naacl/DevlinCLT19} and RoBERTa~\cite{DBLP:journals/corr/abs-1907-11692} compute contextualized embeddings based on the context, leading to superior performance in various downstream tasks. 
Even for the same token, the contextualized embeddings vary for sentences, creating a distribution in the embedding space.

Research has been done to explore the relationship between word frequency and contextualized embeddings. 
\citet{DBLP:conf/acl/Wannasuphoprasit23} showed a correlation between the frequency of a word and the mean norm of its BERT embeddings.
\citet{DBLP:conf/icann/LiangCZRG21} found a negative correlation between the frequency and the norm of BERT embeddings.
\citet{DBLP:journals/corr/abs-2104-08465,DBLP:conf/acl/ZhouECJ22} observed that higher frequency words tend to have a larger radius of the smallest enclosing sphere of their BERT embeddings. 
In particular, the larger radius value means the broader distribution of the embeddings. 
These studies reveal intriguing relationships between word frequency, the norm of embeddings, and the spread of their distribution.

\begin{figure}[t!]
    \centering
    \includegraphics[width=\columnwidth]{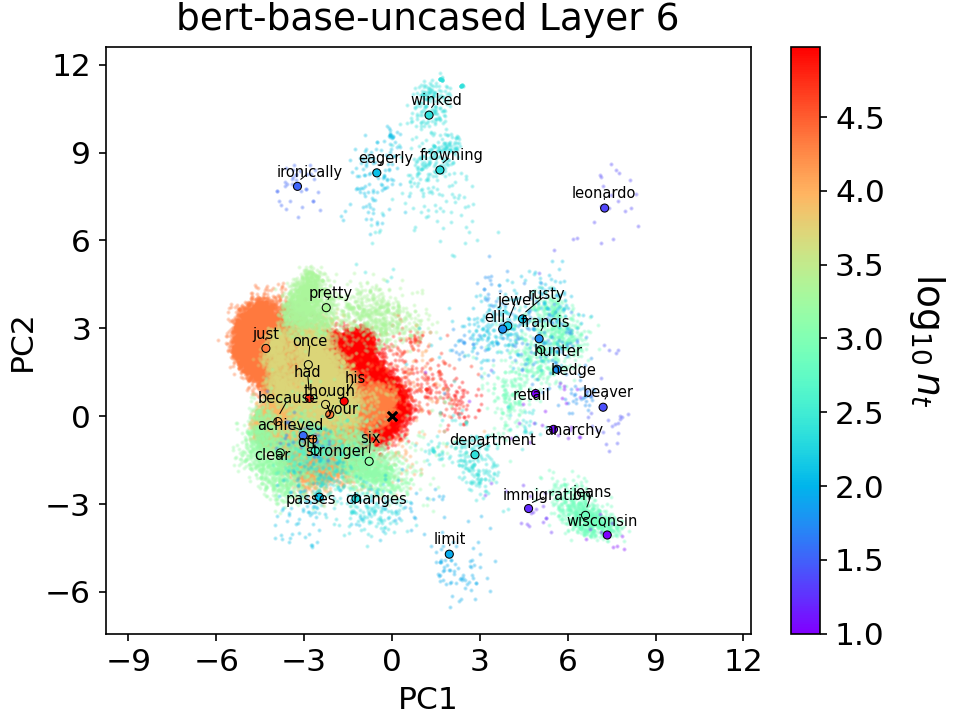}
    \caption{
Scatter plots of PCA-transformed embeddings for the embedding sets $X_t$ of selected tokens.
The origin is indicated by $\times$.
Tokens distributed near the origin exhibit larger variance, whereas tokens farther from the origin exhibit smaller variance.
Embeddings are colored according to token frequency $n_t$.
}
\label{fig:intro}
\end{figure}

\begin{figure*}[t!]
    \centering
    \includegraphics[width=\textwidth]{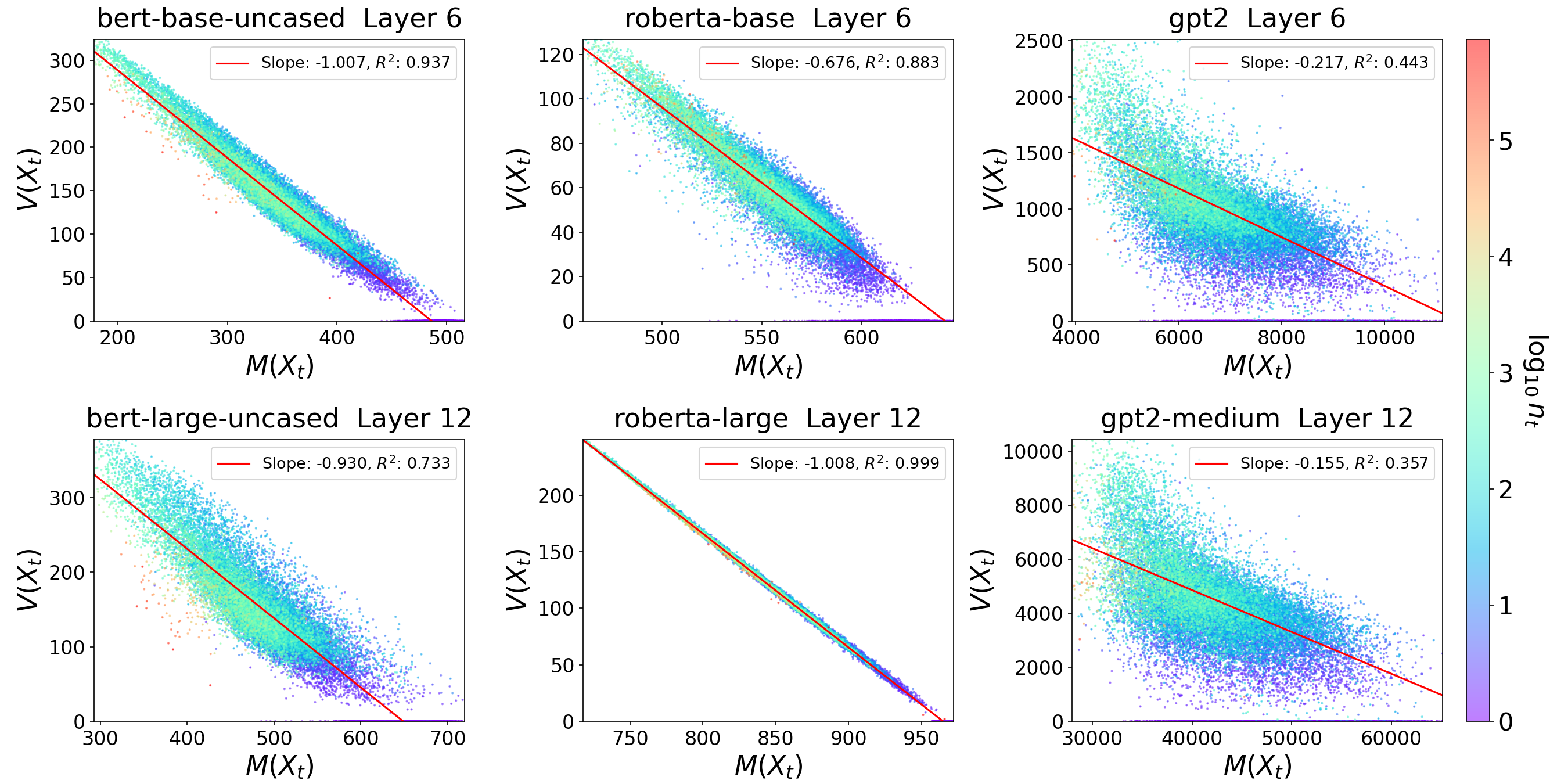}
    \caption{
Scatter plots of $V(X_t)$ against $M(X_t)$ for the middle-layer embeddings of six models with regression lines, slopes, and coefficients of determination, $R^2$.
A consistent trade-off between $M(X_t)$ and $V(X_t)$ is observed in the intermediate layer of each model.
A summary for all the other layers can be found in Fig.~\ref{fig:MV_slope_and_score}. 
Only tokens with $1 \leq \log_{10} n_t \leq 5$ were used for regressions to reduce the influence of extreme values.
}
\label{fig:MV}
\end{figure*}

Based on these existing studies, we analyze the distribution of embeddings using statistical measures computed from the first and second moments of the embedding components. 
Let $\bm{x}_{t,i}$ denote the $d$-dimensional contextualized embedding for token type $t$ in its $i$-th occurrence\footnote{Hereafter, we simply refer to ``token type'' as ``token'' for brevity.}.
For the set of contextualized embeddings $X_t = \{ \bm{x}_{t,1}, \bm{x}_{t,2},\ldots\} \subset \mathbb{R}^d$ corresponding to token $t$, we focus on three values: \textbf{the mean squared norm} $Q(X_t)$, \textbf{the squared norm of the mean embedding} $M(X_t)$, and \textbf{the sum of the variances of each component} $V(X_t)$.
In particular, since the norm of an embedding represents the strength of its meaning~\cite{DBLP:conf/emnlp/OyamaYS23}, $M(X_t)$ represents the strength of the meaning of the token $t$, while $V(X_t)$ can be interpreted as the spread of the distribution based on the variance.

In this paper, we focus on the following identity involving these three values:
\begin{align}
Q(X_t) = M(X_t) + V(X_t). \label{eq:QMVXt}
\end{align}
As can be seen by rewriting this equation as $V(X_t) = Q(X_t) - M(X_t)$, this is nothing more than the well-known formula for variance in elementary statistics.
Furthermore, we experimentally demonstrate that the variation of $Q(X_t)$ from the embeddings of intermediate layers in various Transformer models is small.
Therefore, according to (\ref{eq:QMVXt}), $M(X_t)$ and $V(X_t)$ exhibit a strong trade-off relationship: when the meaning of a token is weaker, the variance of its embeddings is larger, whereas when the meaning is stronger, the variance is smaller.

To observe the trade-off relationship between $M(X_t)$ and $V(X_t)$, Fig.~\ref{fig:intro} shows PCA-transformed embeddings derived from the 6th layer of \texttt{bert-base-uncased}. 
We sampled tokens with frequencies evenly distributed in the range from $10^1$ to $10^5$ for visualization purposes (see Appendix~\ref{app:intro} for more details). 
Tokens whose embeddings are distributed near the origin tend to have a mean embedding closer to the origin, resulting in smaller $M(X_t)$ and larger $V(X_t)$, whereas tokens whose embeddings are distributed farther from the origin have larger $M(X_t)$ and smaller $V(X_t)$.
For example, the tokens \emph{once} and \emph{winked} have similar $Q(X_t)$ values of $494.1$ and $485.6$, respectively. 
However, the embedding set for \emph{once} is closer to the origin than that for \emph{winked}, with $M(X_t)$ values of $239.9$ for \emph{once} and $404.5$ for \emph{winked}. 
Conversely, the variance $V(X_t)$ for \emph{once} is $254.2$, larger than $81.1$ for \emph{winked}.
These results are consistent with the fact that \emph{once} functions as a stopword\footnote{\emph{once} is included in the stopword list provided by NLTK~\cite{DBLP:conf/acl/Bird06}.} with minimal semantic content.

To examine whether the trade-off relationship between $M(X_t)$ and $V(X_t)$, observed in Fig.~\ref{fig:intro}, holds across the intermediate layers of other Transformer models, Fig.~\ref{fig:MV} presents scatter plots of $M(X_t)$ and $V(X_t)$ for the middle-layer embeddings of six models. Consistently, the variation in $Q(X_t)$, which represents the sum of $M(X_t)$ and $V(X_t)$, remains small, confirming the trade-off relationship between $M(X_t)$ and $V(X_t)$.
A detailed layer-wise analysis of this trade-off relationship is provided in Section~\ref{sec:experiments}.

We have obtained interesting insights not only into the set of embeddings for each token, $X_t$, but also into the set of embeddings for all tokens combined, $X \subset \mathbb{R}^d$.
In addition to the identity similar to (\ref{eq:QMVXt}), $Q(X) = M(X) + V(X)$, we focus on the decomposition formula for variance, $V(X) = V_W(X) + V_B(X)$, where $V_W(X)$ is the within-group variance, and $V_B(X)$ is the between-group variance. Through the experiments in Section~\ref{sec:experiments} using these values, we demonstrate that the embeddings in the deeper layers of Transformer models exhibit greater anisotropy.

Our main contributions are as follows:
\begin{itemize}
    \item We focus on three statistical measures, $Q(X_t)$, $M(X_t)$, and $V(X_t)$, to analyze the distribution of contextualized embeddings. We derive the relationship in (\ref{eq:QMVXt}) and introduce an efficient method for computing $V(X_t)$ sequentially.
    \item We experimentally demonstrate that the variation of $Q(X_t)$ is small for embeddings from intermediate layers of various models, and that $M(X_t)$ and $V(X_t)$ exhibit a strong trade-off relationship.
    We theoretically argue that the Layer Normalization (LN) in BERT and RoBERTa reduces the variation of $Q(X_t)$.
    \item For the entire embedding set $X$, we derive relationships between $Q(X)$, $M(X)$, $V_W(X)$, and $V_B(X)$. We experimentally show that the layer-wise changes in these values across various Transformer models align well with previous research that highlights the anisotropy of embedding spaces.
\end{itemize}

\section{Related work}
\subsection{Relationship between frequency and contextualized embeddings}\label{sec:freq_rel}
There are three studies related to our work that deal with the relationship between word frequency and contextualized embeddings. 
The first is by \citet{DBLP:conf/acl/Wannasuphoprasit23}, who found that the mean norm of BERT embeddings for the same word correlated with its frequency and proposed a frequency-considered similarity measure. 
In place of the mean norm, we use the mean squared norm $Q(X_t)$. 
The second study is by \citet{DBLP:conf/icann/LiangCZRG21}, who demonstrated a negative correlation between word frequency and the norms of BERT embeddings. 
In place of the norm of the embeddings, we use the squared norm of the mean embedding $M(X_t)$. 
The third study is by \citet{DBLP:journals/corr/abs-2104-08465, DBLP:conf/acl/ZhouECJ22}, who observed that the radius of the smallest enclosing sphere for BERT embeddings of high-frequency words tends to be larger. 
In place of the radius, we use the variance $V(X_t)$.

Based on these existing studies, in Section~\ref{sec:freq_exp}, we investigate the relationship of $Q(X_t)$, $M(X_t)$, and $V(X_t)$ against log frequency using the middle-layer embeddings of BERT.

\subsection{Norms of embeddings}
The norm of an embedding is an easily computed value and has been the focus of extensive research. 
The norm of a word embedding is related to the Kullback-Leibler divergence~\cite{DBLP:conf/emnlp/OyamaYS23}, and embeddings of less informative words typically exhibit shorter norms~\cite{DBLP:journals/corr/SchakelW15,DBLP:journals/corr/abs-1805-09209,DBLP:conf/emnlp/KobayashiKYI20,DBLP:conf/emnlp/YokoiTASI20}. 
\citet{DBLP:conf/acl/DemeterKD20} showed theoretically that norms are dominant in the computation of logits in the final layer.
\citet{DBLP:journals/corr/abs-2406-10984} shows norm-derived artifacts in unnormalized embeddings, focusing on the axes of the embeddings.

\subsection{Distribution of embeddings}
The distribution of contextualized embeddings has been studied extensively. 
Contextualized embedding spaces exhibit anisotropy, primarily due to the influence of low-frequency words~\cite{DBLP:conf/acl/YuSK0RY22}. 
Based on these observations, \citet{zhang2024reconsideringtokenembeddingsdefinitions} proposed a method for constructing embeddings that result in an isotropic distribution. 
\citet{kutuzov-etal-2022-contextualized} demonstrated using ELMo~\cite{DBLP:conf/naacl/PetersNIGCLZ18} that embeddings of polysemous words such as \emph{cell} form clusters according to their meanings. 
\citet{DBLP:conf/emnlp/YamagiwaOS23} discovered that the embedding space after a whitened ICA transformation exhibits a spiky shape.

\subsection{Information in layer-wise embeddings}
Research focusing on the information in layer-wise embeddings is important for understanding models. 
\citet{DBLP:conf/emnlp/Ethayarajh19,DBLP:conf/iclr/CaiHB021,DBLP:conf/eacl/GodeyCS24} showed that the anisotropy of the embedding space increases as the layers of models such as BERT and GPT-2 deepen.
\citet{DBLP:conf/naacl/Liu0BPS19} performed probing tasks using embeddings from different layers of ELMo, GPT-2, and BERT to investigate performance differences. 
\citet{DBLP:conf/naacl/HewittM19} showed that the BERT embeddings from the intermediate layers capture information related to the syntax trees of sentences. 
\citet{DBLP:conf/blackboxnlp/FayyazAMMP21} observed stability in the norms of BERT embeddings across layers. 
\citet{heimersheim2023residual} showed that the norm of the residual stream~\cite{elhage2021mathematical} in GPT-2 increases as the layers deepen.
\citet{DBLP:conf/coling/SajjadADD22} showed that the variance of the embeddings differs by layer and proposed an effective post-processing.

\section{Token-wise embedding set \texorpdfstring{$X_t$}{Xt}}\label{sec:Xt}
In this section, we first define the token-wise embedding set, $X_t$, for a given token $t$. 
Next, we provide detailed definitions of the statistical measures $Q(X_t)$, $M(X_t)$, and $V(X_t)$, and explain the relationship in (\ref{eq:QMVXt}). 
Finally, we show that the statistical measures of $X_t$ can be efficiently computed through sequential computation.

\subsection{Definition of \texorpdfstring{$X_t$}{Xt}}\label{sec:defXt}
We provide a formal definition of $X_t$, expanding on the brief explanation introduced in Section~\ref{sec:intro}.
Let $T$ be the set of token types present in the corpus. 
For each token $t\in T$, let $S_t$ be the set of sentences in the corpus that contain the token $t$. 
Given a contextualized embedding model $f$ of dimension $d$, let $f(s,t)\in\mathbb{R}^d$ be the embedding\footnote{When the same token type appears multiple times in a single sentence, embeddings are actually computed for each occurrence separately. However, for simplicity of notation, we present it as if there is a single embedding for the token in the sentence.} of token $t$ in a sentence $s\in S_t$. 
For the token $t$, the set of embeddings derived from $f$ and $S_t$ is defined as:
\begin{align}
    X_t := \left\{f(s,t) \mid s\in S_t\right\} \subset \mathbb{R}^d. \label{eq:Xt}
\end{align}
We define the frequency of token $t$ as $n_t := |X_t|$.

\subsection{Statistical measures for \texorpdfstring{$X_t$}{Xt}}\label{sec:defQXtMXtVXt}
We provide a formal definition of $Q(X_t)$, $M(X_t)$, and $V(X_t)$ for $X_t \subset \mathbb{R}^d$, and explain their relationships.
First, we define the mean embedding as
\begin{align}
    \bm{\mu}(X_t) &:= \mathbb{E}_{\bm{x}\in X_t} \left\{\bm{x}\right\} = \frac{1}{|X_t|}\sum_{\bm{x}\in X_t}\bm{x}\in\mathbb{R}^d,\label{eq:MuXt}
\end{align}
where $\mathbb{E}_{\bm{x}\in X_t} \{\cdot\}$ represents the sample mean over $X_t$. 
Next, for $X_t$, the mean squared norm $Q(X_t)$, the squared norm of the mean embedding $M(X_t)$, and the sum of the variances of each component $V(X_t)$ are defined as follows:
\begin{align}
    Q(X_t) &:= \mathbb{E}_{\bm{x}\in X_t} \left\{ \|\bm{x}\|^2\right\}, \label{eq:QXt}  \\
    M(X_t) &:= \|\mathbb{E}_{\bm{x}\in X_t} \left\{ \bm{x}\right\}\|^2 = \| \bm{\mu}(X_t) \|^2, \label{eq:MXt}  \\
\begin{split}
V(X_t) &:= \mathbb{E}_{\bm{x}\in X_t} \left\{ \|\bm{x} - \bm{\mu} (X_t) \|^2\right\}  \\
&= \sum_{i=1}^d \mathbb{E}_{\bm{x}\in X_t} \left\{ (x_i - \mu_i (X_t))^2\right\}, \label{eq:VXt}
\end{split}
\end{align}
where $x_i$ and $\mu_i(X_t)$ are the $i$-th components of $\bm{x}$ and $\bm{\mu}(X_t)$, respectively, and $\|\cdot\|$ denotes the $L_2$ norm. 
A larger $M(X_t)$ indicates that $X_t$ is farther from the origin. 
Since the norm of an embedding represents the strength of its meaning~\cite{DBLP:conf/emnlp/OyamaYS23}, a larger $M(X_t)$ indicates that token $t$ carries greater semantic content. 
A larger $V(X_t)$ indicates a wider distribution within $X_t$, which suggests greater variability in the meaning of token $t$.
Then, calculations (see Appendix~\ref{app:Xt}) yield the identity
\begin{align*}
    Q(X_t) = M(X_t) + V(X_t),
\end{align*}
which is exactly~(\ref{eq:QMVXt}) in Section~\ref{sec:intro}. 
Thus, $V(X_t)$ can be determined as $Q(X_t) - M(X_t)$, the difference between two norm-derived values.

\begin{algorithm}[t!]
\caption{Sequential computation of $n_t$, $\bm{\mu}(X_t)$, $Q(X_t)$, $M(X_t)$, and $V(X_t)$ for each token $t$}
\label{algo:updating}
\begin{algorithmic}[1]
\Require A contextualized embedding model $f$,
\Statex\hspace{-0.1cm}a corpus $S=\bigcup_{t\in T}S_t$
\Ensure A dictionary $\mathcal{D}$ mapping each token $t$ to 
\Statex\hspace{-0.1cm}$\mathcal{D}[t]=(n_t, \bm{\mu}(X_t), Q(X_t), M(X_t), V(X_t))$
\Statex
\State{Initialize an empty dictionary $\mathcal{D}$}
\For{each sentence $s\in S$}
\For{each token occurrence $t \in s$}
\State // Compute the token embedding
\State $\bm{x}\leftarrow f(s, t)\in\mathbb{R}^d$
\If{the token $t$ is already a key in $\mathcal{D}$}
\State // Load previous values
\State $(k, \bm{u}, q, \_, \_) \leftarrow \mathcal{D}[t]$
\State // Compute new values sequentially
\State $k'\leftarrow k+1$
\State $\bm{u}' \leftarrow \frac{k}{k + 1} \bm{u} + \frac{1}{k + 1}\bm{x}\in\mathbb{R}^d$
\State $q'\leftarrow \frac{k}{k + 1} q + \frac{1}{k + 1}\|\bm{x}\|^2$
\State $m'\leftarrow \|\bm{u}'\|^2$
\State $v'\leftarrow q' - m'$
\State // Update the dictionary
\State $\mathcal{D}[t]\leftarrow (k', \bm{u}', q', m', v')$
\Else
\State // Initialize on first occurrence of $t$
\State $\mathcal{D}[t]\leftarrow (1, \bm{x}, \|\bm{x}\|^2, \|\bm{x}\|^2, 0)$
\EndIf
\EndFor
\EndFor
\end{algorithmic}
\end{algorithm}

\subsection{Efficient computation for \texorpdfstring{$X_t$}{Xt}}\label{sec:calcXt}
Storing all $X_t$ when computing $Q(X_t)$, $M(X_t)$, and $V(X_t)$ is inefficient. 
This inefficiency can be addressed by sequentially computing $Q(X_t)$ and $\bm{\mu}(X_t)$. 
Using the sequentially computed $Q(X_t)$ and $\bm{\mu}(X_t)$, $M(X_t)$ and $V(X_t)$ can also be computed\footnote{Sequential computation methods for variance, such as Welford's online algorithm~\cite{welford1962note}, have been known for a long time.} based on (\ref{eq:MXt}) and (\ref{eq:QMVXt}).
The procedure\footnote{In practice, embeddings are usually computed in batches.} is detailed in Algorithm~\ref{algo:updating}.  
This algorithm requires storing only $|T|$ embeddings for $\bm{\mu}(X_t)$ and $4|T|$ scalar values, allowing for efficient handling of the statistical measures for $X_t$.

\section{The entire embedding set \texorpdfstring{$X$}{X}}\label{sec:X}
In Section~\ref{sec:Xt}, we considered the embedding set $X_t$ for each token. 
Considering the entire embedding set $X$, which includes all embedding sets $X_t$, we can also analyze the entire embedding space. 
Therefore, in this section, we first provide the definition of $X$ and then define the statistical measures for $X$ as we did for $X_t$. 
Furthermore, we show that the total variance $V(X)$ can be decomposed into the within-group variance and the between-group variance.
Finally, we explain the efficient computation for $X$.

\subsection{Definition of \texorpdfstring{$X$}{X}}\label{sec:defX}
With $X_t$, the entire embedding set $X\subset \mathbb{R}^d$ is defined as follows:
\begin{align}
    X := \bigcup_{t\in T}X_t \subset \mathbb{R}^d,  \label{eq:X}
\end{align}
where the number of embeddings in $X$ is defined as $n := |X| = \sum_{t\in T} n_t$.

Replacing $X_t$ in (\ref{eq:MuXt}) with $X$, we can define the mean embedding $\bm{\mu}(X)\in \mathbb{R}^d$ for $X$. 
Similarly, replacing $X_t$ in (\ref{eq:QXt}), (\ref{eq:MXt}), and (\ref{eq:VXt}) with $X$, we can define $Q(X)$, $M(X)$, and $V(X)$, respectively.
A larger $M(X)$ indicates that $\bm{\mu}(X)$ is farther from the origin, making the embedding space more anisotropic. 
A larger $V(X)$ indicates a wider spread within the embedding space.
Replacing $X_t$ with $X$ in (\ref{eq:QMVXt}), the following identity also holds:
\begin{align}
Q(X) = M(X) + V(X).\label{eq:QMVX}
\end{align}

\begin{figure}[t!]
    \centering
    \includegraphics[width=0.98\columnwidth]{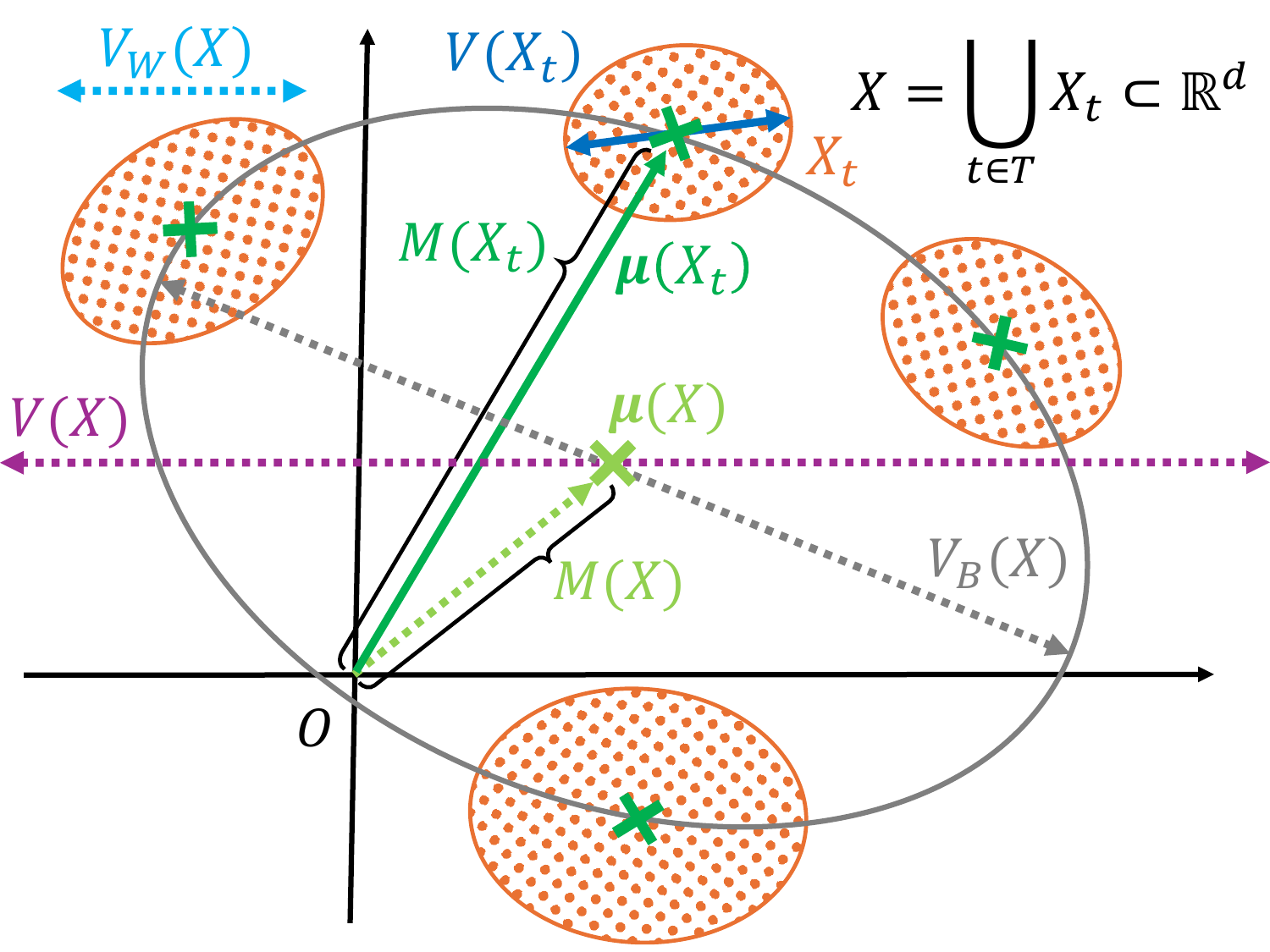}
    \caption{
Illustration of the token-wise embedding sets $X_t$, $t\in T$, and the entire embedding set $X$.
The values $\bm{\mu}(X_t)$, $M(X_t)$, and $V(X_t)$ are computed for each $X_t$, while $\bm{\mu}(X)$, $M(X)$, and $V(X)$ are for $X$.
In addition, $V(X)$ is decomposed into the within-group variance $V_W(X)$ and the between-group variance $V_B(X)$. 
$V_W(X)$ is the frequency-weighted mean of $V(X_t)$, while $V_B(X)$ represents the spread of  $\bm{\mu}(X_t)$ around $\bm{\mu}(X)$. 
Although $M$ and $V$ are illustrated as a norm and a standard deviation, respectively, they are actually the squared versions as shown in (\ref{eq:MXt}) and (\ref{eq:VXt}).
}
\label{fig:within_between}
\end{figure}

\subsection{Within-group variance and between-group variance}
In general, variance can be decomposed into within-group variance, which represents the spread within clusters, and between-group variance, which represents the spread between clusters~\cite{muthen1991multilevel}. 
Accordingly, by treating $\{X_t\}_{t\in T}$ as clusters, we consider the decomposition of the variance $V(X)$ of the entire embedding set $X$ into the within-group variance $V_W(X)$ and the between-group variance $V_B(X)$ as follows:
\begin{align}
V(X) = V_W(X) + V_B(X). \label{eq:VwVb}
\end{align}
In the context of clustering, these can also be referred to as the within-cluster variance and the between-cluster variance, respectively.
Calculations (see Appendix~\ref{app:VwVb}) show that:
\begin{align}
V_W(X) &= \sum_{t\in T}p_t V(X_t), \label{eq:Vw} \\
V_B(X) &= \sum_{t \in T} p_t \|\bm{\mu}(X_t) - \bm{\mu}(X)\|^2, \label{eq:Vb}
\end{align}
where
\begin{align}
p_t := |X_t|/|X| = n_t/n.\label{eq:pt}
\end{align}
Thus, $V_W(X)$ is the frequency-weighted mean of $V(X_t)$ and indicates the spread within each $X_t$. 
On the other hand, $V_B(X)$ is the frequency-weighted mean of $\|\bm{\mu}(X_t) - \bm{\mu}(X)\|^2$ and indicates the spread between $\bm{\mu}(X_t)$ around $\bm{\mu}(X)$.

From (\ref{eq:QMVX}) and (\ref{eq:VwVb}), the values $Q(X)$, $M(X)$, $V_W(X)$, and $V_B(X)$ satisfy:
\begin{align}
Q(X) = M(X) + V_W(X) + V_B(X). \label{eq:QMVwVb}
\end{align}
Figure~\ref{fig:within_between} illustrates the relationships among these values computed from $X_t$ and $X$. 
While the values for $X_t$ are computed for each token, the values for $X$ are computed from the entire embedding space.

\subsection{Efficient computation for \texorpdfstring{$X$}{X}}\label{sec:calcX}
In Section~\ref{sec:calcXt}, we showed that the statistical measures for $X_t$ can be computed efficiently using a sequential method.
Similarly, the statistical measures for $X$ can also be computed efficiently by using the statistical measures for $X_t$.

As seen in Section~\ref{sec:defX}, $n$ can be obtained as the sum of $n_t$. 
Additionally, simple calculations (see Appendix~\ref{app:calcX}) show that $\bm{\mu}(X)$ and $Q(X)$ can be expressed as the frequency-weighted means of $\bm{\mu}(X_t)$ and $Q(X_t)$, respectively:
\begin{align}
\bm{\mu}(X)&=\mathbb{E}_{\bm{x}\in X} \left\{\bm{x}\right\}=\sum_{t\in T} p_t\bm{\mu}(X_t),\label{eq:MuX}\\
Q(X)&=\mathbb{E}_{\bm{x}\in X} \left\{ \|\bm{x}\|^2\right\} = \sum_{t\in T} p_t Q(X_t).\label{eq:QX}
\end{align}
These expressions enable efficient computation of the statistical measures for $X$.
Furthermore, using these values, $M(X)$, $V(X)$, $V_W(X)$, and $V_B(X)$ can also be computed efficiently.

\section{Experiments}\label{sec:experiments}
In this section, we conduct experiments using contextualized embedding models to calculate the statitical measures for $X_t$ and $X$ as described in Sections~\ref{sec:Xt} and~\ref{sec:X}.
First, we explain the experimental settings, and then present the results for $X_t$ and $X$.
Note that in this study, we focus on token embeddings instead of word embeddings\footnote{This is because we found artifacts in the experimental results when representing a word embedding as the mean of the token embeddings. For details, refer to Appendix~\ref{app:wordartifact}.}, and we do not distinguish between whether a token corresponds to a complete word or a part of a word\footnote{For example, in BERT tokenization, both the \emph{ing} token and the \emph{\#\#ing} token are treated the same as the \emph{ing} token.}.

\begin{table}[t!]
\centering
\begin{tabular}{@{\hspace{0.2em}}l@{\hspace{0.5em}}r@{\hspace{0.5em}}r@{\hspace{0.5em}}r@{\hspace{0.2em}}}
\toprule
Model & Layers & Dims. & Params.\\
\midrule
\texttt{bert-base-uncased} & \multirow{3}{*}{13} & \multirow{3}{*}{768} & 110M \\
\texttt{roberta-base} &  &  & 125M\\
\texttt{gpt2} &  & & 117M \\
\midrule
\texttt{bert-large-uncased} &  \multirow{3}{*}{25} &  \multirow{3}{*}{1024} & 340M \\
\texttt{roberta-large} & & & 355M \\
\texttt{gpt2-medium} & & & 345M \\
\bottomrule
\end{tabular}
\caption{
The number of layers including the input layer, the dimensions, and the parameter size for each model.
}
\label{tab:model_layer_dim}
\end{table}

\begin{figure*}[t!]
    \centering
    \includegraphics[width=\textwidth]{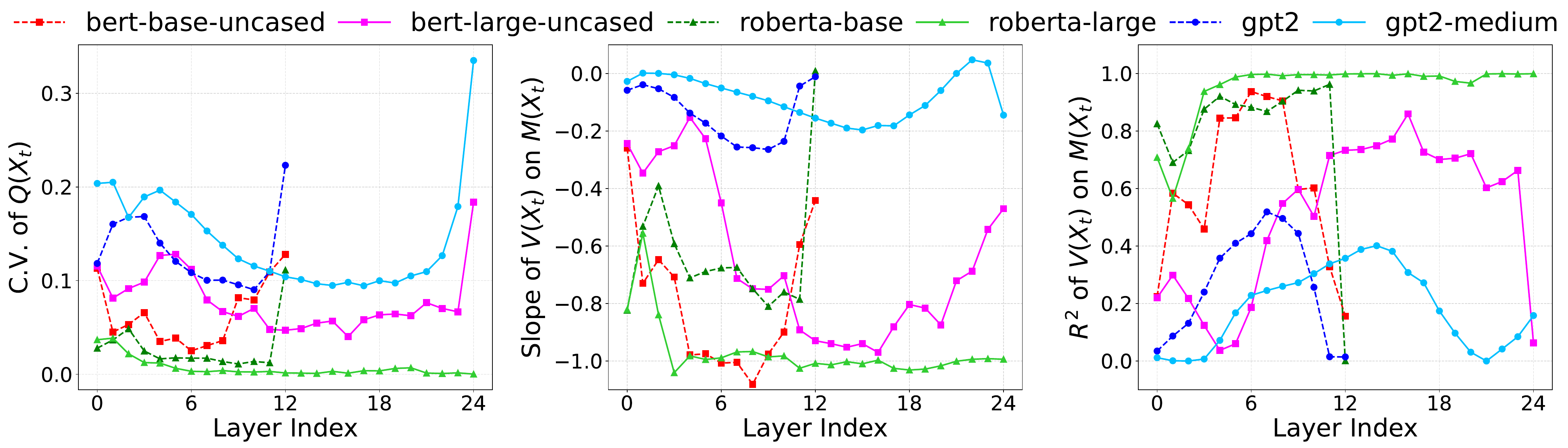}
    \caption{
For each layer across the six models, the coefficient of variation (C.V.) of $Q(X_t)$ on the left, the slope of the regression line of $V(X_t)$ on $M(X_t)$ in the middle, and the corresponding coefficient of determination $R^2$ on the right are shown. 
For all models, the C.V. approximately reaches its minimum in the intermediate layers. 
Consequently, the slope and $R^2$ approximately reach their minimum and maximum, respectively, in the intermediate layers.
Only tokens with $1 \leq \log_{10} n_t \leq 5$ were used to reduce the influence of extreme values.
}
\label{fig:MV_slope_and_score}
\end{figure*}

\subsection{Settings}
\subsubsection{Models}
We used the \texttt{transformers} library~\cite{DBLP:conf/emnlp/WolfDSCDMCRLFDS20} in our experiments.
Following \citet{DBLP:conf/icann/LiangCZRG21,DBLP:conf/acl/ZhouECJ22,DBLP:conf/acl/Wannasuphoprasit23}, we used the BERT~\cite{DBLP:conf/naacl/DevlinCLT19} models \texttt{bert-base-uncased} and \texttt{bert-large-uncased}. 
Additionally, we also used the RoBERTa~\cite{DBLP:journals/corr/abs-1907-11692} models \texttt{roberta-base} and \texttt{roberta-large}, and the GPT-2~\cite{radford2019language} models \texttt{gpt2} and \texttt{gpt2-medium}. 
The number of layers, the dimensions, and the size of the parameters for each model are shown in Table~\ref{tab:model_layer_dim}.

\subsubsection{Dataset}\label{sec:datasets}
Similar to \citet{DBLP:conf/acl/Wannasuphoprasit23}, we used the BookCorpus~\cite{DBLP:conf/iccv/ZhuKZSUTF15}. 
For efficiency, we randomly sampled $1\%$ of the sentences from the corpus and selected those containing fewer than 64 words for the embedding computations.
The total number of sampled sentences was $739{,}106$. 
Details of the number of tokens, $|T|\approx 24\textrm{k}$, and the number of embeddings, $|X|\approx 12\textrm{M}$, are provided in Table~\ref{tab:T_N} in Appendix~\ref{app:datasets}.
The histograms of sentence lengths and the frequency of $\log_{10} n_t$ are also shown in Figs.~\ref{fig:sent_hist} and \ref{fig:token_count}, respectively, in Appendix~\ref{app:datasets}.

\subsection{Results for the token-wise embedding sets}\label{sec:expXt}
Figure~\ref{fig:MV} shows scatter plots of $V(X_t)$ against $M(X_t)$ from the middle-layer embeddings of the six models. 
Each scatter plot shows the regression line and displays its slope and the coefficient of determination, $R^2$. 
Consistently, the sum of $M(X_t)$ and $V(X_t)$, namely $Q(X_t)$, exhibits small variation, confirming the trade-off relationship between $M(X_t)$ and $V(X_t)$. 
Furthermore, the slopes of the regression lines are negative, with large $R^2$ values. 
For example, in the case of \texttt{roberta-large}, the slope of the regression line is $-1.008$ and $R^2 = 0.999$, indicating a nearly perfect trade-off relationship with a constant sum.

Next, we examine the variation of $Q(X_t)$ and the trade-off between $M(X_t)$ and $V(X_t)$ across layers. 
Figure~\ref{fig:MV_slope_and_score} shows the coefficient of variation\footnote{C.V. is defined as the ratio of the standard deviation to the mean, representing the relative variability in the data.} (C.V.) of $Q(X_t)$, the slope of the regression line of $V(X_t)$ on $M(X_t)$, and the corresponding $R^2$ value for each layer of the six models.
The C.V. of $Q(X_t)$ is generally low and it reaches its minimum value approximately in the intermediate layers of each model, where the trade-off between $M(X_t)$ and $V(X_t)$ becomes  more pronounced.
In the intermediate layers of BERT and RoBERTa, the slope of the regression line reaches a minimum value of approximately $-1$, and the $R^2$ value approaches its maximum of $1$.
However, in the case of GPT-2, the minimum C.V. of $Q(X_t)$ is larger than those of BERT and RoBERTa, with a minimum slope of approximately $-0.2$ and a maximum $R^2$ value of around $0.5$. 
These differences are likely due to architectural differences, which will be discussed in Section~\ref{sec:discussion}.

\begin{figure*}[t!]
    \centering
    \includegraphics[width=\textwidth]{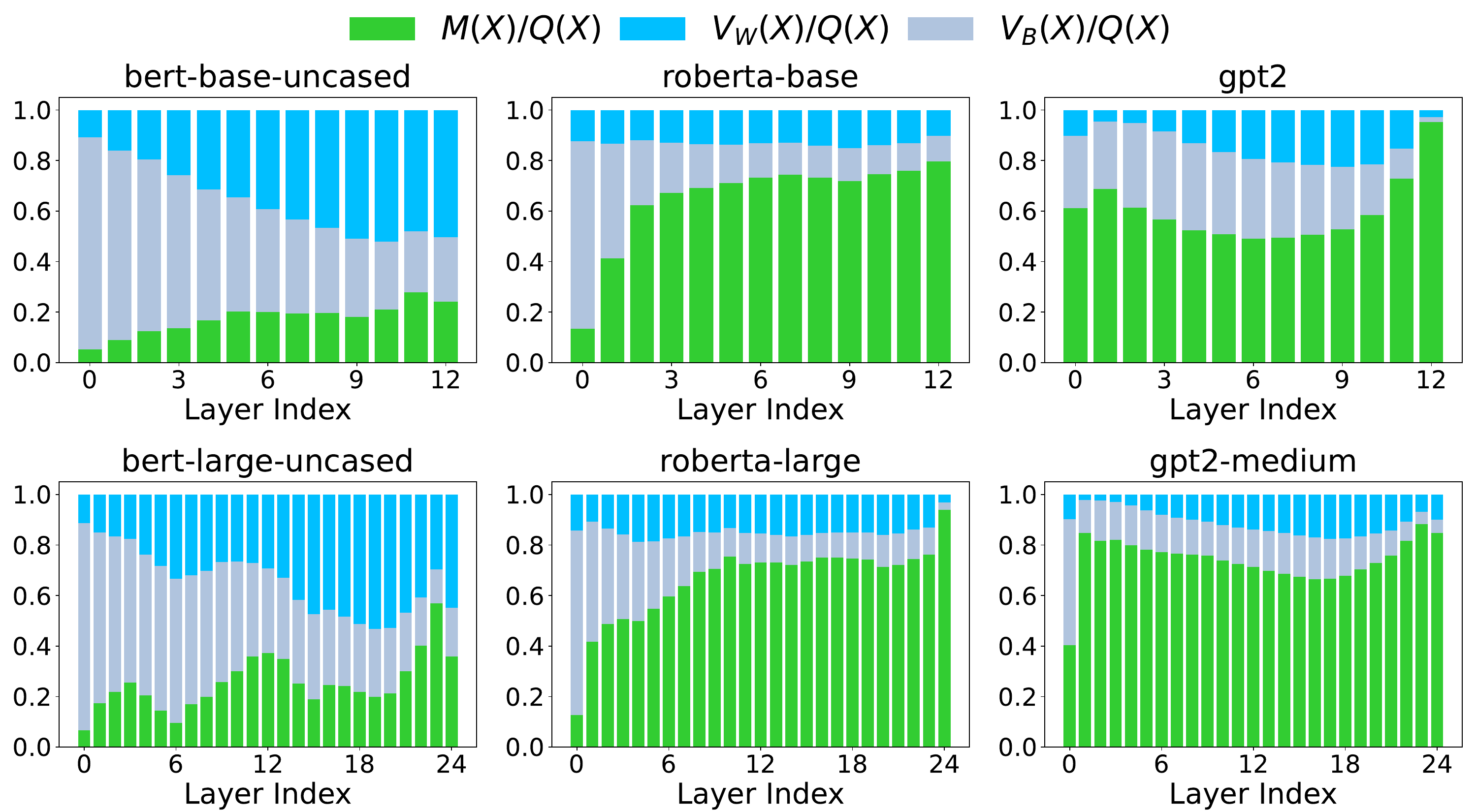}
    \caption{
The ratios of $M(X)$, $V_W(X)$, and $V_B(X)$, each normalized by $Q(X)$, for each layer across the six models.  
As the layers deepen, the ratio of $M(X)$ tends to exceed that of $V_W(X) + V_B(X) (=V(X))$.  
Meanwhile, the ratio of $V_W(X)$ increases relative to $V_B(X)$.  
Figure~\ref{fig:Vw_per_V} shows detailed comparisons between $V_W(X)$ and $V_B(X)$.
Further plots of the ratios of these values and those of the original values are shown in Figs.~\ref{fig:MX_VwX_VbX_per_QX} and \ref{fig:QX_MX_VX}, respectively, in Appendix~\ref{app:X}.
Only tokens with $1 \leq \log_{10} n_t \leq 5$ were used to reduce the influence of extreme values.}
\label{fig:MV_bar}
\end{figure*}

\begin{figure}[t!]
    \centering
    \includegraphics[width=\columnwidth]{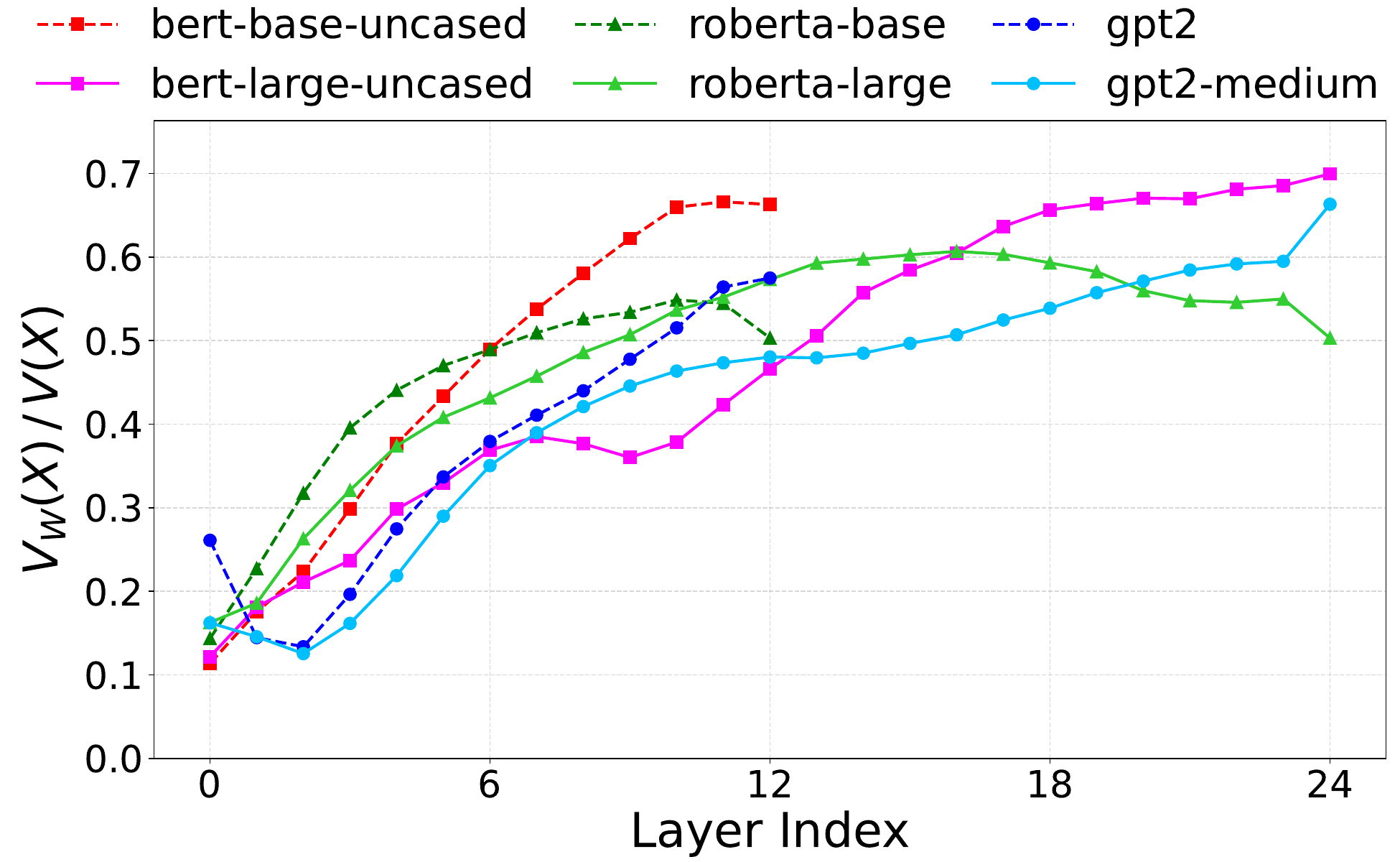}
    \caption{
The ratio $V_W(X)/V(X)$ in Fig.~\ref{fig:MV_bar} for each layer across the six models.
As the layers deepen, the ratio of $V_W(X)$ increases.
Further plots of these values are shown in Fig.~\ref{fig:Vw_Vb_Vb-per-V} in Appendix~\ref{app:X}.
}
\label{fig:Vw_per_V}
\end{figure}

\subsection{Results for the entire embedding set}\label{sec:expX}
As seen in (\ref{eq:QMVwVb}), $Q(X)$ can be decomposed into $M(X)$, $V_W(X)$, and $V_B(X)$. 
Figure~\ref{fig:MV_bar} illustrates the changes in the ratios of $M(X)$, $V_W(X)$, and $V_B(X)$ normalized by $Q(X)$ across the layers of the six models.
Generally, as the layers deepen, the ratio of $M(X)$ increases, which means that the ratio of the sum $V_W(X) + V_B(X)$ (equal to $V(X)$) decreases. 
Additionally, a comparison between $V_W(X)$ and $V_B(X)$ shows that the ratio of $V_W(X)$ increases as the layers deepen.
Figure~\ref{fig:QX_MX_VX} in Appendix~\ref{app:X} presents the original layer-wise values of $Q(X)$, $M(X)$, and $V(X)$.

According to (\ref{eq:VwVb}), $V_W(X)+V_B(X)=V(X)$.
To investigate the value of $V_W(X)$ relative to $V_B(X)$, Fig.~\ref{fig:Vw_per_V} shows the ratio $V_W(X)/V(X)$.
Consistent with the results in Fig.~\ref{fig:MV_bar}, the ratio of $V_W(X)$ increases in each model as the layers deepen, i.e., the ratio of $V_B(X)$ decreases.

Previous studies on the anisotropy of embedding spaces across layers~\cite{DBLP:conf/emnlp/Ethayarajh19,DBLP:conf/iclr/CaiHB021,DBLP:conf/eacl/GodeyCS24} showed that for BERT, RoBERTa, and GPT-2, the average cosine similarity between randomly sampled words increases as the layers deepen.  
This finding is consistent with our results in Fig.~\ref{fig:MV_bar}, where the ratio of $M(X)$ increases and the ratio of $V(X)$ decreases as the layers deepen, and in Fig.~\ref{fig:Vw_per_V}, where the ratio of $V_B(X)$ decreases.  
These studies also found that the cosine similarity between embeddings of the same word in different sentences decreases as the layers deepen.  
This observation is also consistent with our results in Fig.~\ref{fig:Vw_per_V}, where the ratio of $V_W(X)$ gradually increases.  
While previous work such as \citet{DBLP:conf/emnlp/Ethayarajh19} computed cosine similarities by randomly sampling $1{,}000$ embeddings, we computed the values using all embeddings in the dataset.

\begin{figure*}[t!]
    \centering
    \includegraphics[width=\textwidth]{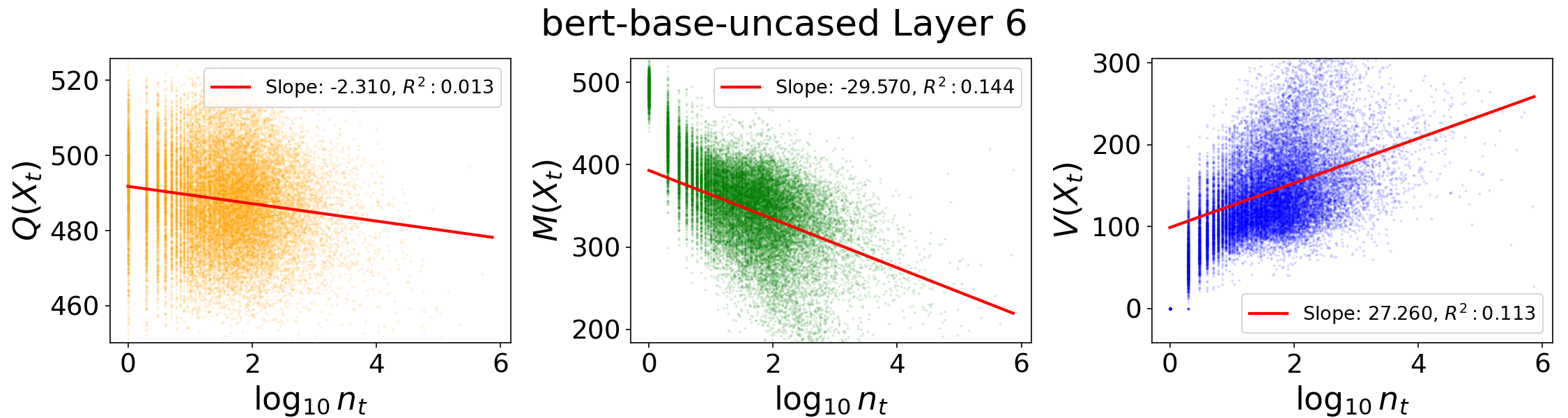}
    \caption{
Scatter plots of (left) $Q(X_t)$, (middle) $M(X_t)$, and (right) $V(X_t)$ against $\log_{10} n_t$ for the middle-layer embeddings of \texttt{bert-base-uncased}.
Each plot includes a regression line, its slope, and the coefficient of determination ($R^2$).
While the slope of $Q(X_t)$ is close to zero, those of $M(X_t)$ and $V(X_t)$ are negative and positive, respectively.
Only tokens with $1 \leq \log_{10} n_t \leq 5$ were used for regressions to reduce the influence of extreme values.
Appendix~\ref{app:QMV_MV} presents the results for embeddings from multiple layers of several models based on log-scaled values.
}
\label{fig:freq_vs_QXt-MXt-VXt}
\end{figure*}

\subsection{Relationship of \texorpdfstring{$Q(X_t)$}{Q(Xt)}, \texorpdfstring{$M(X_t)$}{M(Xt)}, and \texorpdfstring{$V(X_t)$}{V(Xt)} against token frequency}\label{sec:freq_exp}
In Section~\ref{sec:freq_rel}, we discussed three related studies that examined the relationship between word frequency and values associated with $Q(X_t)$, $M(X_t)$, and $V(X_t)$.
In this section, we examine the correlations between these three proposed values and token frequency.

Figure~\ref{fig:freq_vs_QXt-MXt-VXt} presents scatter plots of $Q(X_t)$, $M(X_t)$, and $V(X_t)$ against log frequency, using embeddings from the 6th layer of \texttt{bert-base-uncased} as a representative example.
The slope of $Q(X_t)$ remains stable and close to zero. 
In contrast, the negative slope of $M(X_t)$ and the positive slope of $V(X_t)$ indirectly suggest a trade-off relationship between $M(X_t)$ and $V(X_t)$.
Similar trends were observed across different layers and models (see Appendix~\ref{app:QMV_MV}).

\section{Discussion}\label{sec:discussion}
\subsection{Why does the C.V. of \texorpdfstring{$Q(X_t)$}{Q(Xt)} reach its minimum in the intermediate layers?}
In this study, we discovered that the C.V. of $Q(X_t)$ decreases in the intermediate layers of various Transformer models, as shown in Fig.~\ref{fig:MV_slope_and_score}.
While further investigation into the reasons behind this phenomenon remains as future work, we present our hypothesis here.

In the input layer (layer 0), pre-trained embeddings are used, whereas in the final layer, embeddings are influenced by the objective function and the computation of logits. 
As a result, embeddings in these two layers are expected to exhibit different characteristics compared to those in other layers. 
Indeed, as shown in Fig.~\ref{fig:QX_MX_VX} in Appendix~\ref{app:X}, the plots of $Q(X)$, $M(X)$, and $V(X)$ for each layer indicate that the values in the input and output layers differ significantly from those in other layers. 
This suggests that the influence of these specialized layers is reduced in the intermediate layers, possibly reflecting a property inherent to the model architecture or language --- namely, the reduced variation in $Q(X_t)$.

\subsection{Why does GPT-2 behave differently from BERT and RoBERTa in Fig.~\ref{fig:MV_slope_and_score}?}
In Fig.~\ref{fig:MV_slope_and_score}, although GPT-2 shows similar trends to BERT and RoBERTa, the minimum C.V. of $Q(X_t)$, the minimum slope of the regression line, and the maximum $R^2$ differ from those of BERT and RoBERTa.
Furthermore, for GPT-2, the values of $Q(X_t)$, $M(X_t)$, and $V(X_t)$ in Fig.~\ref{fig:MV}, as well as $Q(X)$, $M(X)$, and $V(X)$ in Fig.~\ref{fig:QX_MX_VX} in Appendix~\ref{app:X}, differ significantly in magnitude from those for BERT and RoBERTa.

This is likely due to the different placement of layer normalization (LN) within the Transformer layer. 
In general, as shown in Fig.~\ref{fig:layer_diff} in Appendix~\ref{app:LN}, Post-LN Transformers such as BERT and RoBERTa apply the LN after the feed-forward network (FF), while Pre-LN Transformers such as GPT-2 apply it before the FF~\cite{DBLP:conf/icml/XiongYHZZXZLWL20}.
Consequently, the embeddings $\bm{x}\in X_t$ of BERT and RoBERTa are outputs of the LN, and the squared norm $\|\bm{x}\|^2$ is controlled with small variation.

For Post-LN Transformers, under certain specific conditions assuming an ideal scenario, it can be shown that the C.V. of $Q(X_t)$ is sufficiently small.
As shown in Appendix~\ref{app:theory-layer-nomalization},
\begin{align}
C.V.(Q(X_t)) = O\left(\frac{1}{\sqrt{n_0}}\right),\label{eq:CVQ}
\end{align}
where $n_0 = \min_{t\in T}n_t$. 
This result indicates that the C.V. approaches zero as all $n_t$ increase.

On the other hand,  $\|\bm{x}\|^2$ for GPT-2 is not controlled by the LN, yet it is interesting to observe similar trends in Figs.~\ref{fig:MV_bar} and~\ref{fig:Vw_per_V}.
Although it is not necessarily desirable for embeddings to be artificially constrained directly by LN, as in BERT and RoBERTa, the trade-off between $M(X_t)$ and $V(X_t)$ is also observed in the GPT-2 model, which is not directly constrained in this way.
This observation suggests that the constraints imposed by LN reflect reality to some extent and did not cause a significant issue in the model's language learning process.

\section{Conclusion}
In this study, we focused on the distribution of contextualized embeddings and analyzed three values: the mean squared norm $Q(X_t)$, the squared norm of the mean embedding $M(X_t)$, and the sum of the variances of each component $V(X_t)$. 
In Section~\ref{sec:Xt}, we showed that the values of $Q(X_t)$, $M(X_t)$, and $V(X_t)$ are related by (\ref{eq:QMVXt}) and can be efficiently computed using a sequential method.
We also found that, in the intermediate layers of several models, the variation of $Q(X_t)$ is small, which results in a strong trade-off between $M(X_t)$ and $V(X_t)$.
We explained in Section~\ref{sec:discussion} that the small variation in $Q(X_t)$ can be attributed to the placement of LN.
The values of $Q(X_t)$, $M(X_t)$, and $V(X_t)$ can also be applied to the entire embedding set $X$, and we demonstrated that the total variance $V(X)$ can be decomposed into within-group variance $V_W(X)$ and between-group variance $V_B(X)$.
As seen in Figs.~\ref{fig:MV_bar} and~\ref{fig:Vw_per_V}, the experimental results from relative comparisons show that as the layers deepen, $M(X)$ increases, while $V(X)$ and $V_B(X)$ decrease, and $V_W(X)$ increases.
These results are consistent with existing studies on the anisotropy of embedding spaces across layers.

\clearpage
\section*{Limitations}
\begin{itemize}
    \item Due to computational resource limitations, we used relatively small models with parameter sizes fewer than 1B, as shown in Table~\ref{tab:model_layer_dim}. 
    Since anisotropy in embedding spaces is affected by parameter size~\cite{godey2024why}, verification with larger models would be desirable. 
    Note that previous studies on the relationship between word frequency and embeddings~\cite{DBLP:conf/icann/LiangCZRG21,DBLP:conf/acl/ZhouECJ22,DBLP:conf/acl/Wannasuphoprasit23} have only examined the final layer of BERT models. 
    In contrast, we conducted experiments with BERT, RoBERTa, and GPT-2, following the settings from previous work on layer-wise anisotropy~\cite{DBLP:conf/emnlp/Ethayarajh19,DBLP:conf/iclr/CaiHB021,DBLP:conf/eacl/GodeyCS24}.
    \item In order to run the experiments efficiently, we did not use the full BookCorpus. The number of sentences used in the experiments was $739{,}106$, and the total number of embeddings $|X|$ exceeded 10 million, which we considered sufficient.
    \item This study deals only with English models. The analysis of the values of $Q$, $M$, and $V$ for different languages using multilingual models is left for future work.
    \item In this study, we analyzed high-dimensional distributions using scalar values such as norms and the sum of variances, prioritizing ease of interpretation, as discussed in Sections~\ref{sec:Xt} and~\ref{sec:X}.
    \item  \citet{DBLP:conf/emnlp/Ethayarajh19,DBLP:conf/iclr/CaiHB021,DBLP:conf/eacl/GodeyCS24} used cosine similarity to examine layer-wise anisotropy, which facilitates comparisons across models and layers. 
    In contrast, the values of $Q$, $M$, and $V$ are not normalized, and these values vary significantly across models and layers. Therefore, appropriate adjustments may be necessary for such comparisons. 
    Based on this, in Figs.~\ref{fig:MV_bar} and~\ref{fig:Vw_per_V}, we have normalized the values using $Q(X)$ and $V(X)$ to allow comparisons across models and layers. 
    This normalization shows, for example, that while the value of $V(X)$ increases as the layers deepen in GPT-2 (Fig.~\ref{fig:QX_MX_VX} in Appendix~\ref{app:X}), the ratio of $V(X)/Q(X)$ decreases, as shown in Fig.~\ref{fig:MV_bar}.
    \item In our experiments, only tokens with $1 \leq \log_{10} n_t \leq 5$ were used when computing values such as the regression line, to reduce the influence of extreme values. The choice of this frequency range is ad hoc, and the influence of token frequency on the results has not been examined in detail.
    \item The probability distribution settings assumed in the theory of the C.V. of $Q(X_t)$ (Appendix~\ref{app:theory-layer-nomalization}) do not necessarily reflect reality, and the derived formulas have only limited value.
    \item A more detailed experimental and theoretical analysis is needed to understand why the C.V. of $Q(X_t)$ becomes smaller in the intermediate layers and why GPT-2 behaves differently from BERT and RoBERTa in Fig.~\ref{fig:MV_slope_and_score}.
\end{itemize}

\section*{Ethics Statement}
This study complies with the \href{https://www.aclweb.org/portal/content/acl-code-ethics}{ACL Ethics Policy}.

\section*{Acknowledgments}
We would like to thank Yusuke Takase, Momose Oyama, Ryo Kishino, and the anonymous reviewers for their helpful comments and suggestions.
This study was partially supported by JSPS KAKENHI 22H05106, 23H03355, JST CREST JPMJCR21N3, JST SPRING JPMJSP2110.

\bibliography{custom}

\begin{thebibliography}{40}
\providecommand{\natexlab}[1]{#1}

\bibitem[{Arefyev et~al.(2018)Arefyev, Ermolaev, and Panchenko}]{DBLP:journals/corr/abs-1805-09209}
Nikolay Arefyev, Pavel Ermolaev, and Alexander Panchenko. 2018.
\newblock \href {https://arxiv.org/abs/1805.09209} {How much does a word weigh? weighting word embeddings for word sense induction}.
\newblock \emph{CoRR}, abs/1805.09209.

\bibitem[{Bird(2006)}]{DBLP:conf/acl/Bird06}
Steven Bird. 2006.
\newblock \href {https://doi.org/10.3115/1225403.1225421} {{NLTK:} the natural language toolkit}.
\newblock In \emph{{ACL} 2006, 21st International Conference on Computational Linguistics and 44th Annual Meeting of the Association for Computational Linguistics, Proceedings of the Conference, Sydney, Australia, 17-21 July 2006}. The Association for Computer Linguistics.

\bibitem[{Cai et~al.(2021)Cai, Huang, Bian, and Church}]{DBLP:conf/iclr/CaiHB021}
Xingyu Cai, Jiaji Huang, Yuchen Bian, and Kenneth Church. 2021.
\newblock \href {https://openreview.net/forum?id=xYGNO86OWDH} {Isotropy in the contextual embedding space: Clusters and manifolds}.
\newblock In \emph{9th International Conference on Learning Representations, {ICLR} 2021, Virtual Event, Austria, May 3-7, 2021}. OpenReview.net.

\bibitem[{Demeter et~al.(2020)Demeter, Kimmel, and Downey}]{DBLP:conf/acl/DemeterKD20}
David Demeter, Gregory Kimmel, and Doug Downey. 2020.
\newblock \href {https://doi.org/10.18653/V1/2020.ACL-MAIN.198} {Stolen probability: {A} structural weakness of neural language models}.
\newblock In \emph{Proceedings of the 58th Annual Meeting of the Association for Computational Linguistics, {ACL} 2020, Online, July 5-10, 2020}, pages 2191--2197. Association for Computational Linguistics.

\bibitem[{Devlin et~al.(2019)Devlin, Chang, Lee, and Toutanova}]{DBLP:conf/naacl/DevlinCLT19}
Jacob Devlin, Ming{-}Wei Chang, Kenton Lee, and Kristina Toutanova. 2019.
\newblock \href {https://doi.org/10.18653/V1/N19-1423} {{BERT:} pre-training of deep bidirectional transformers for language understanding}.
\newblock In \emph{Proceedings of the 2019 Conference of the North American Chapter of the Association for Computational Linguistics: Human Language Technologies, {NAACL-HLT} 2019, Minneapolis, MN, USA, June 2-7, 2019, Volume 1 (Long and Short Papers)}, pages 4171--4186. Association for Computational Linguistics.

\bibitem[{Elhage et~al.(2021)Elhage, Nanda, Olsson, Henighan, Joseph, Mann, Askell, Bai, Chen, Conerly et~al.}]{elhage2021mathematical}
Nelson Elhage, Neel Nanda, Catherine Olsson, Tom Henighan, Nicholas Joseph, Ben Mann, Amanda Askell, Yuntao Bai, Anna Chen, Tom Conerly, et~al. 2021.
\newblock \href {https://transformer-circuits.pub/2021/framework/index.html} {A mathematical framework for transformer circuits}.
\newblock \emph{Transformer Circuits Thread}.

\bibitem[{Ethayarajh(2019)}]{DBLP:conf/emnlp/Ethayarajh19}
Kawin Ethayarajh. 2019.
\newblock \href {https://doi.org/10.18653/V1/D19-1006} {How contextual are contextualized word representations? comparing the geometry of bert, elmo, and {GPT-2} embeddings}.
\newblock In \emph{Proceedings of the 2019 Conference on Empirical Methods in Natural Language Processing and the 9th International Joint Conference on Natural Language Processing, {EMNLP-IJCNLP} 2019, Hong Kong, China, November 3-7, 2019}, pages 55--65. Association for Computational Linguistics.

\bibitem[{Fayyaz et~al.(2021)Fayyaz, Aghazadeh, Modarressi, Mohebbi, and Pilehvar}]{DBLP:conf/blackboxnlp/FayyazAMMP21}
Mohsen Fayyaz, Ehsan Aghazadeh, Ali Modarressi, Hosein Mohebbi, and Mohammad~Taher Pilehvar. 2021.
\newblock \href {https://doi.org/10.18653/V1/2021.BLACKBOXNLP-1.29} {Not all models localize linguistic knowledge in the same place: {A} layer-wise probing on bertoids' representations}.
\newblock In \emph{Proceedings of the Fourth BlackboxNLP Workshop on Analyzing and Interpreting Neural Networks for NLP, BlackboxNLP@EMNLP 2021, Punta Cana, Dominican Republic, November 11, 2021}, pages 375--388. Association for Computational Linguistics.

\bibitem[{Godey et~al.(2024{\natexlab{a}})Godey, de~la Clergerie, and Sagot}]{DBLP:conf/eacl/GodeyCS24}
Nathan Godey, {\'{E}}ric~Villemonte de~la Clergerie, and Beno{\^{\i}}t Sagot. 2024{\natexlab{a}}.
\newblock \href {https://aclanthology.org/2024.eacl-long.3} {Anisotropy is inherent to self-attention in transformers}.
\newblock In \emph{Proceedings of the 18th Conference of the European Chapter of the Association for Computational Linguistics, {EACL} 2024 - Volume 1: Long Papers, St. Julian's, Malta, March 17-22, 2024}, pages 35--48. Association for Computational Linguistics.

\bibitem[{Godey et~al.(2024{\natexlab{b}})Godey, de~la Clergerie, and Sagot}]{godey2024why}
Nathan Godey, {\'E}ric~Villemonte de~la Clergerie, and Beno{\^\i}t Sagot. 2024{\natexlab{b}}.
\newblock \href {https://openreview.net/forum?id=MoitXWlXcS} {Why do small language models underperform? studying language model saturation via the softmax bottleneck}.
\newblock In \emph{First Conference on Language Modeling}.

\bibitem[{Heimersheim and Turner(2023)}]{heimersheim2023residual}
Stefan Heimersheim and Alex Turner. 2023.
\newblock \href {https://www.lesswrong.com/posts/8mizBCm3dyc432nK8/residual-stream-norms-grow-exponentially-over-the-forward} {Residual stream norms grow exponentially over the forward pass}.
\newblock \emph{LESSWRONG}.

\bibitem[{Hewitt and Manning(2019)}]{DBLP:conf/naacl/HewittM19}
John Hewitt and Christopher~D. Manning. 2019.
\newblock \href {https://doi.org/10.18653/V1/N19-1419} {A structural probe for finding syntax in word representations}.
\newblock In \emph{Proceedings of the 2019 Conference of the North American Chapter of the Association for Computational Linguistics: Human Language Technologies, {NAACL-HLT} 2019, Minneapolis, MN, USA, June 2-7, 2019, Volume 1 (Long and Short Papers)}, pages 4129--4138. Association for Computational Linguistics.

\bibitem[{Kobayashi et~al.(2020)Kobayashi, Kuribayashi, Yokoi, and Inui}]{DBLP:conf/emnlp/KobayashiKYI20}
Goro Kobayashi, Tatsuki Kuribayashi, Sho Yokoi, and Kentaro Inui. 2020.
\newblock \href {https://doi.org/10.18653/V1/2020.EMNLP-MAIN.574} {Attention is not only a weight: Analyzing transformers with vector norms}.
\newblock In \emph{Proceedings of the 2020 Conference on Empirical Methods in Natural Language Processing, {EMNLP} 2020, Online, November 16-20, 2020}, pages 7057--7075. Association for Computational Linguistics.

\bibitem[{Kobayashi et~al.(2024)Kobayashi, Kuribayashi, Yokoi, and Inui}]{DBLP:conf/iclr/KobayashiKYI24}
Goro Kobayashi, Tatsuki Kuribayashi, Sho Yokoi, and Kentaro Inui. 2024.
\newblock \href {https://openreview.net/forum?id=mYWsyTuiRp} {Analyzing feed-forward blocks in transformers through the lens of attention maps}.
\newblock In \emph{The Twelfth International Conference on Learning Representations, {ICLR} 2024, Vienna, Austria, May 7-11, 2024}. OpenReview.net.

\bibitem[{Kutuzov et~al.(2022)Kutuzov, Velldal, and {\O}vrelid}]{kutuzov-etal-2022-contextualized}
Andrey Kutuzov, Erik Velldal, and Lilja {\O}vrelid. 2022.
\newblock \href {https://doi.org/10.3384/nejlt.2000-1533.2022.3478} {Contextualized embeddings for semantic change detection: Lessons learned}.
\newblock In \emph{Northern European Journal of Language Technology, Volume 8}, Copenhagen, Denmark. Northern European Association of Language Technology.

\bibitem[{Liang et~al.(2021)Liang, Cao, Zheng, Ren, and Gao}]{DBLP:conf/icann/LiangCZRG21}
Yuxin Liang, Rui Cao, Jie Zheng, Jie Ren, and Ling Gao. 2021.
\newblock \href {https://doi.org/10.1007/978-3-030-86383-8\_36} {Learning to remove: Towards isotropic pre-trained {BERT} embedding}.
\newblock In \emph{Artificial Neural Networks and Machine Learning - {ICANN} 2021 - 30th International Conference on Artificial Neural Networks, Bratislava, Slovakia, September 14-17, 2021, Proceedings, Part {V}}, volume 12895 of \emph{Lecture Notes in Computer Science}, pages 448--459. Springer.

\bibitem[{Liu et~al.(2019{\natexlab{a}})Liu, Gardner, Belinkov, Peters, and Smith}]{DBLP:conf/naacl/Liu0BPS19}
Nelson~F. Liu, Matt Gardner, Yonatan Belinkov, Matthew~E. Peters, and Noah~A. Smith. 2019{\natexlab{a}}.
\newblock \href {https://doi.org/10.18653/V1/N19-1112} {Linguistic knowledge and transferability of contextual representations}.
\newblock In \emph{Proceedings of the 2019 Conference of the North American Chapter of the Association for Computational Linguistics: Human Language Technologies, {NAACL-HLT} 2019, Minneapolis, MN, USA, June 2-7, 2019, Volume 1 (Long and Short Papers)}, pages 1073--1094. Association for Computational Linguistics.

\bibitem[{Liu et~al.(2019{\natexlab{b}})Liu, Ott, Goyal, Du, Joshi, Chen, Levy, Lewis, Zettlemoyer, and Stoyanov}]{DBLP:journals/corr/abs-1907-11692}
Yinhan Liu, Myle Ott, Naman Goyal, Jingfei Du, Mandar Joshi, Danqi Chen, Omer Levy, Mike Lewis, Luke Zettlemoyer, and Veselin Stoyanov. 2019{\natexlab{b}}.
\newblock \href {https://arxiv.org/abs/1907.11692} {Roberta: {A} robustly optimized {BERT} pretraining approach}.
\newblock \emph{CoRR}, abs/1907.11692.

\bibitem[{Mikolov et~al.(2013)Mikolov, Sutskever, Chen, Corrado, and Dean}]{DBLP:conf/nips/MikolovSCCD13}
Tom{\'{a}}s Mikolov, Ilya Sutskever, Kai Chen, Gregory~S. Corrado, and Jeffrey Dean. 2013.
\newblock \href {https://proceedings.neurips.cc/paper/2013/hash/9aa42b31882ec039965f3c4923ce901b-Abstract.html} {Distributed representations of words and phrases and their compositionality}.
\newblock In \emph{Advances in Neural Information Processing Systems 26: 27th Annual Conference on Neural Information Processing Systems 2013. Proceedings of a meeting held December 5-8, 2013, Lake Tahoe, Nevada, United States}, pages 3111--3119.

\bibitem[{Muth{\'e}n(1991)}]{muthen1991multilevel}
Bengt~O Muth{\'e}n. 1991.
\newblock Multilevel factor analysis of class and student achievement components.
\newblock \emph{Journal of Educational measurement}, 28(4):338--354.

\bibitem[{Oyama et~al.(2023)Oyama, Yokoi, and Shimodaira}]{DBLP:conf/emnlp/OyamaYS23}
Momose Oyama, Sho Yokoi, and Hidetoshi Shimodaira. 2023.
\newblock \href {https://doi.org/10.18653/V1/2023.EMNLP-MAIN.131} {Norm of word embedding encodes information gain}.
\newblock In \emph{Proceedings of the 2023 Conference on Empirical Methods in Natural Language Processing, {EMNLP} 2023, Singapore, December 6-10, 2023}, pages 2108--2130. Association for Computational Linguistics.

\bibitem[{Pennington et~al.(2014)Pennington, Socher, and Manning}]{DBLP:conf/emnlp/PenningtonSM14}
Jeffrey Pennington, Richard Socher, and Christopher~D. Manning. 2014.
\newblock \href {https://doi.org/10.3115/V1/D14-1162} {Glove: Global vectors for word representation}.
\newblock In \emph{Proceedings of the 2014 Conference on Empirical Methods in Natural Language Processing, {EMNLP} 2014, October 25-29, 2014, Doha, Qatar, {A} meeting of SIGDAT, a Special Interest Group of the {ACL}}, pages 1532--1543. {ACL}.

\bibitem[{Peters et~al.(2018)Peters, Neumann, Iyyer, Gardner, Clark, Lee, and Zettlemoyer}]{DBLP:conf/naacl/PetersNIGCLZ18}
Matthew~E. Peters, Mark Neumann, Mohit Iyyer, Matt Gardner, Christopher Clark, Kenton Lee, and Luke Zettlemoyer. 2018.
\newblock \href {https://doi.org/10.18653/V1/N18-1202} {Deep contextualized word representations}.
\newblock In \emph{Proceedings of the 2018 Conference of the North American Chapter of the Association for Computational Linguistics: Human Language Technologies, {NAACL-HLT} 2018, New Orleans, Louisiana, USA, June 1-6, 2018, Volume 1 (Long Papers)}, pages 2227--2237. Association for Computational Linguistics.

\bibitem[{Radford et~al.(2019)Radford, Wu, Child, Luan, Amodei, and Sutskever}]{radford2019language}
Alec Radford, Jeff Wu, Rewon Child, David Luan, Dario Amodei, and Ilya Sutskever. 2019.
\newblock Language models are unsupervised multitask learners.

\bibitem[{Sajjad et~al.(2022)Sajjad, Alam, Dalvi, and Durrani}]{DBLP:conf/coling/SajjadADD22}
Hassan Sajjad, Firoj Alam, Fahim Dalvi, and Nadir Durrani. 2022.
\newblock \href {https://aclanthology.org/2022.coling-1.277} {Effect of post-processing on contextualized word representations}.
\newblock In \emph{Proceedings of the 29th International Conference on Computational Linguistics, {COLING} 2022, Gyeongju, Republic of Korea, October 12-17, 2022}, pages 3127--3142. International Committee on Computational Linguistics.

\bibitem[{Schakel and Wilson(2015)}]{DBLP:journals/corr/SchakelW15}
Adriaan M.~J. Schakel and Benjamin~J. Wilson. 2015.
\newblock \href {https://arxiv.org/abs/1508.02297} {Measuring word significance using distributed representations of words}.
\newblock \emph{CoRR}, abs/1508.02297.

\bibitem[{Vaswani et~al.(2017)Vaswani, Shazeer, Parmar, Uszkoreit, Jones, Gomez, Kaiser, and Polosukhin}]{DBLP:conf/nips/VaswaniSPUJGKP17}
Ashish Vaswani, Noam Shazeer, Niki Parmar, Jakob Uszkoreit, Llion Jones, Aidan~N. Gomez, Lukasz Kaiser, and Illia Polosukhin. 2017.
\newblock \href {https://proceedings.neurips.cc/paper/2017/hash/3f5ee243547dee91fbd053c1c4a845aa-Abstract.html} {Attention is all you need}.
\newblock In \emph{Advances in Neural Information Processing Systems 30: Annual Conference on Neural Information Processing Systems 2017, December 4-9, 2017, Long Beach, CA, {USA}}, pages 5998--6008.

\bibitem[{Wannasuphoprasit et~al.(2023)Wannasuphoprasit, Zhou, and Bollegala}]{DBLP:conf/acl/Wannasuphoprasit23}
Saeth Wannasuphoprasit, Yi~Zhou, and Danushka Bollegala. 2023.
\newblock \href {https://doi.org/10.18653/V1/2023.FINDINGS-ACL.550} {Solving cosine similarity underestimation between high frequency words by $\ell_2$ norm discounting}.
\newblock In \emph{Findings of the Association for Computational Linguistics: {ACL} 2023, Toronto, Canada, July 9-14, 2023}, pages 8644--8652. Association for Computational Linguistics.

\bibitem[{Welford(1962)}]{welford1962note}
Barry~Payne Welford. 1962.
\newblock Note on a method for calculating corrected sums of squares and products.
\newblock \emph{Technometrics}, 4(3):419--420.

\bibitem[{Wolf et~al.(2020)Wolf, Debut, Sanh, Chaumond, Delangue, Moi, Cistac, Rault, Louf, Funtowicz, Davison, Shleifer, von Platen, Ma, Jernite, Plu, Xu, Scao, Gugger, Drame, Lhoest, and Rush}]{DBLP:conf/emnlp/WolfDSCDMCRLFDS20}
Thomas Wolf, Lysandre Debut, Victor Sanh, Julien Chaumond, Clement Delangue, Anthony Moi, Pierric Cistac, Tim Rault, R{\'{e}}mi Louf, Morgan Funtowicz, Joe Davison, Sam Shleifer, Patrick von Platen, Clara Ma, Yacine Jernite, Julien Plu, Canwen Xu, Teven~Le Scao, Sylvain Gugger, Mariama Drame, Quentin Lhoest, and Alexander~M. Rush. 2020.
\newblock \href {https://doi.org/10.18653/V1/2020.EMNLP-DEMOS.6} {Transformers: State-of-the-art natural language processing}.
\newblock In \emph{Proceedings of the 2020 Conference on Empirical Methods in Natural Language Processing: System Demonstrations, {EMNLP} 2020 - Demos, Online, November 16-20, 2020}, pages 38--45. Association for Computational Linguistics.

\bibitem[{Xiong et~al.(2020)Xiong, Yang, He, Zheng, Zheng, Xing, Zhang, Lan, Wang, and Liu}]{DBLP:conf/icml/XiongYHZZXZLWL20}
Ruibin Xiong, Yunchang Yang, Di~He, Kai Zheng, Shuxin Zheng, Chen Xing, Huishuai Zhang, Yanyan Lan, Liwei Wang, and Tie{-}Yan Liu. 2020.
\newblock \href {http://proceedings.mlr.press/v119/xiong20b.html} {On layer normalization in the transformer architecture}.
\newblock In \emph{Proceedings of the 37th International Conference on Machine Learning, {ICML} 2020, 13-18 July 2020, Virtual Event}, volume 119 of \emph{Proceedings of Machine Learning Research}, pages 10524--10533. {PMLR}.

\bibitem[{Yamagiwa et~al.(2023)Yamagiwa, Oyama, and Shimodaira}]{DBLP:conf/emnlp/YamagiwaOS23}
Hiroaki Yamagiwa, Momose Oyama, and Hidetoshi Shimodaira. 2023.
\newblock \href {https://doi.org/10.18653/V1/2023.EMNLP-MAIN.283} {Discovering universal geometry in embeddings with {ICA}}.
\newblock In \emph{Proceedings of the 2023 Conference on Empirical Methods in Natural Language Processing, {EMNLP} 2023, Singapore, December 6-10, 2023}, pages 4647--4675. Association for Computational Linguistics.

\bibitem[{Yamagiwa et~al.(2024)Yamagiwa, Oyama, and Shimodaira}]{DBLP:journals/corr/abs-2406-10984}
Hiroaki Yamagiwa, Momose Oyama, and Hidetoshi Shimodaira. 2024.
\newblock \href {https://doi.org/10.48550/ARXIV.2406.10984} {Revisiting cosine similarity via normalized ica-transformed embeddings}.
\newblock \emph{CoRR}, abs/2406.10984.

\bibitem[{Yokoi et~al.(2020)Yokoi, Takahashi, Akama, Suzuki, and Inui}]{DBLP:conf/emnlp/YokoiTASI20}
Sho Yokoi, Ryo Takahashi, Reina Akama, Jun Suzuki, and Kentaro Inui. 2020.
\newblock \href {https://doi.org/10.18653/V1/2020.EMNLP-MAIN.236} {Word rotator's distance}.
\newblock In \emph{Proceedings of the 2020 Conference on Empirical Methods in Natural Language Processing, {EMNLP} 2020, Online, November 16-20, 2020}, pages 2944--2960. Association for Computational Linguistics.

\bibitem[{Yu et~al.(2022)Yu, Song, Kim, Lee, Ryu, and Yoon}]{DBLP:conf/acl/YuSK0RY22}
Sangwon Yu, Jongyoon Song, Heeseung Kim, Seongmin Lee, Woo{-}Jong Ryu, and Sungroh Yoon. 2022.
\newblock \href {https://doi.org/10.18653/V1/2022.ACL-LONG.3} {Rare tokens degenerate all tokens: Improving neural text generation via adaptive gradient gating for rare token embeddings}.
\newblock In \emph{Proceedings of the 60th Annual Meeting of the Association for Computational Linguistics (Volume 1: Long Papers), {ACL} 2022, Dublin, Ireland, May 22-27, 2022}, pages 29--45. Association for Computational Linguistics.

\bibitem[{Zhang et~al.(2024)Zhang, Li, and Okumura}]{zhang2024reconsideringtokenembeddingsdefinitions}
Ying Zhang, Dongyuan Li, and Manabu Okumura. 2024.
\newblock \href {https://arxiv.org/abs/2408.01308} {Reconsidering token embeddings with the definitions for pre-trained language models}.
\newblock \emph{Preprint}, arXiv:2408.01308.

\bibitem[{Zhou et~al.(2022{\natexlab{a}})Zhou, Ethayarajh, Card, and Jurafsky}]{DBLP:conf/acl/ZhouECJ22}
Kaitlyn Zhou, Kawin Ethayarajh, Dallas Card, and Dan Jurafsky. 2022{\natexlab{a}}.
\newblock \href {https://doi.org/10.18653/V1/2022.ACL-SHORT.45} {Problems with cosine as a measure of embedding similarity for high frequency words}.
\newblock In \emph{Proceedings of the 60th Annual Meeting of the Association for Computational Linguistics (Volume 2: Short Papers), {ACL} 2022, Dublin, Ireland, May 22-27, 2022}, pages 401--423. Association for Computational Linguistics.

\bibitem[{Zhou et~al.(2021)Zhou, Ethayarajh, and Jurafsky}]{DBLP:journals/corr/abs-2104-08465}
Kaitlyn Zhou, Kawin Ethayarajh, and Dan Jurafsky. 2021.
\newblock \href {https://arxiv.org/abs/2104.08465} {Frequency-based distortions in contextualized word embeddings}.
\newblock \emph{CoRR}, abs/2104.08465.

\bibitem[{Zhou et~al.(2022{\natexlab{b}})Zhou, Ethayarajh, and Jurafsky}]{DBLP:conf/acl/ZhouEJ22}
Kaitlyn Zhou, Kawin Ethayarajh, and Dan Jurafsky. 2022{\natexlab{b}}.
\newblock \href {https://doi.org/10.18653/V1/2022.FINDINGS-ACL.164} {Richer countries and richer representations}.
\newblock In \emph{Findings of the Association for Computational Linguistics: {ACL} 2022, Dublin, Ireland, May 22-27, 2022}, pages 2074--2085. Association for Computational Linguistics.

\bibitem[{Zhu et~al.(2015)Zhu, Kiros, Zemel, Salakhutdinov, Urtasun, Torralba, and Fidler}]{DBLP:conf/iccv/ZhuKZSUTF15}
Yukun Zhu, Ryan Kiros, Richard~S. Zemel, Ruslan Salakhutdinov, Raquel Urtasun, Antonio Torralba, and Sanja Fidler. 2015.
\newblock \href {https://doi.org/10.1109/ICCV.2015.11} {Aligning books and movies: Towards story-like visual explanations by watching movies and reading books}.
\newblock In \emph{2015 {IEEE} International Conference on Computer Vision, {ICCV} 2015, Santiago, Chile, December 7-13, 2015}, pages 19--27. {IEEE} Computer Society.

\end{thebibliography}

\appendix
\onecolumn

\section{Details of Fig.~\ref{fig:intro}}\label{app:intro}

\begin{figure}[t!]
    \centering
    \includegraphics[height=7cm, keepaspectratio]{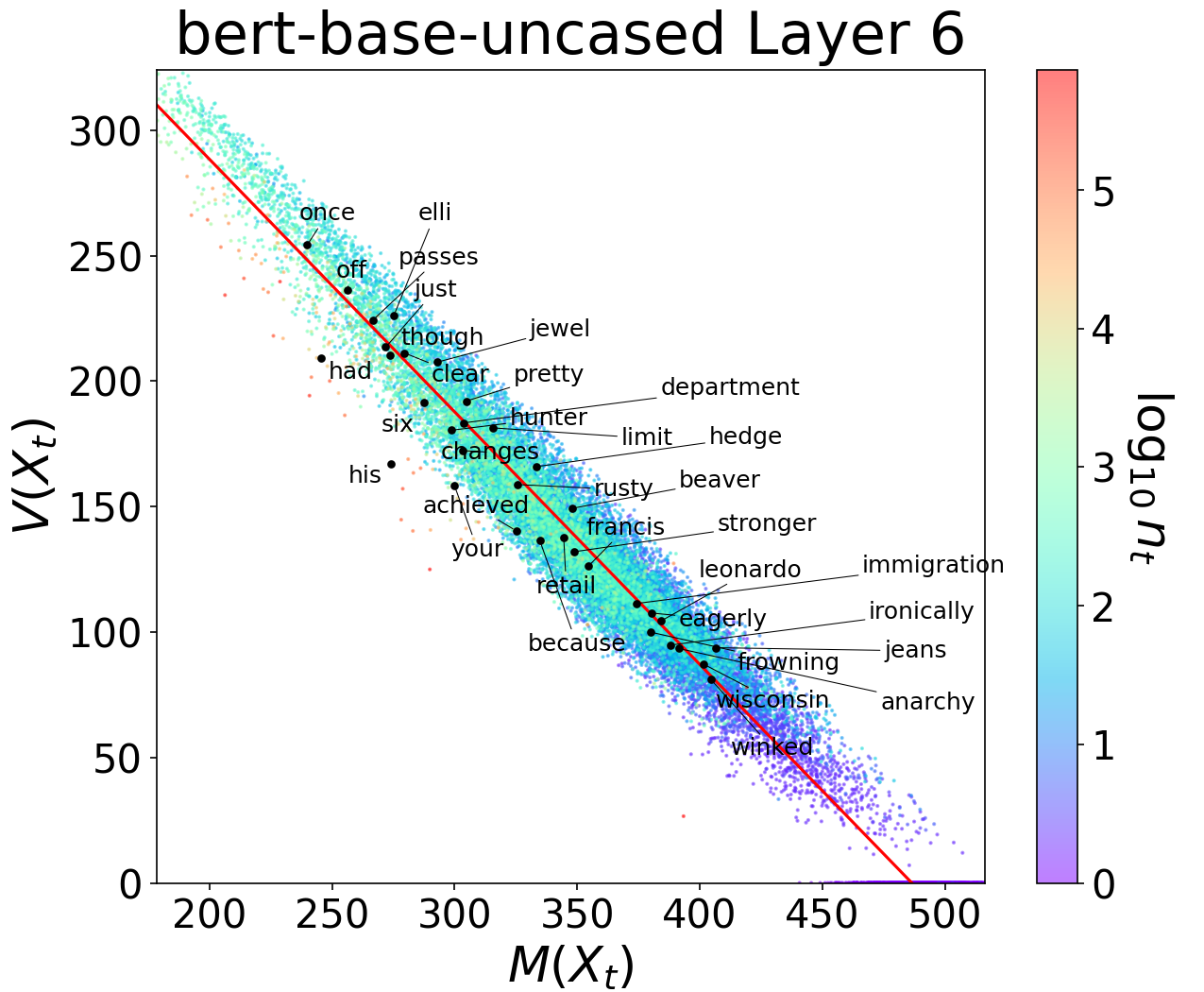}
    \caption{
Scatter plots of $M(X_t)$ and $V(X_t)$ using the 6th layer of \texttt{bert-base-uncased}, along with the regression line and selected tokens from Fig.~\ref{fig:intro}. The tokens are close to the regression line, indicating that the tokens have similar $Q(X_t)$ values.
}
\label{fig:MV_for_intro}
\end{figure}

\begin{table}[t!]
\small
\centering
\begin{tabular}{lrrrr}
\toprule
token $t$ & $n_t$ & $Q(X_t)$ & $M(X_t)$ & $V(X_t)$ \\
\midrule
\emph{his} & 94718 & 441.0 & 273.9 & 167.1 \\
\emph{had} & 56623 & 454.8 & 245.7 & 209.1 \\
\emph{just} & 22619 & 485.4 & 271.9 & 213.5 \\
\emph{your} & 20307 & 458.2 & 299.9 & 158.3 \\
\emph{off} & 12005 & 492.7 & 256.3 & 236.4 \\
\emph{because} & 8484 & 471.2 & 334.7 & 136.5 \\
\emph{though} & 5085 & 484.1 & 273.7 & 210.4 \\
\emph{once} & 5022 & 494.1 & 239.9 & 254.2 \\
\emph{pretty} & 2082 & 496.7 & 304.7 & 192.0 \\
\emph{clear} & 1479 & 490.7 & 279.4 & 211.3 \\
\emph{six} & 1302 & 479.2 & 287.6 & 191.6 \\
\emph{hunter} & 770 & 478.9 & 298.5 & 180.4 \\
\emph{jeans} & 727 & 500.3 & 406.6 & 93.7 \\
\emph{stronger} & 435 & 480.6 & 348.6 & 132.0 \\
\emph{department} & 233 & 487.0 & 303.7 & 183.3 \\
\emph{winked} & 229 & 485.6 & 404.5 & 81.1 \\
\emph{changes} & 214 & 475.7 & 303.4 & 172.3 \\
\emph{frowning} & 210 & 479.8 & 379.8 & 100.0 \\
\emph{jewel} & 157 & 500.2 & 292.7 & 207.5 \\
\emph{rusty} & 124 & 484.3 & 325.5 & 158.8 \\
\emph{eagerly} & 122 & 488.0 & 380.4 & 107.6 \\
\emph{passes} & 118 & 490.9 & 266.6 & 224.3 \\
\emph{limit} & 88 & 496.9 & 315.5 & 181.4 \\
\emph{elli} & 61 & 501.3 & 275.2 & 226.1 \\
\emph{hedge} & 59 & 499.4 & 333.5 & 165.9 \\
\emph{francis} & 57 & 480.8 & 354.5 & 126.3 \\
\emph{achieved} & 35 & 465.4 & 325.1 & 140.3 \\
\emph{ironically} & 33 & 483.0 & 388.2 & 94.8 \\
\emph{beaver} & 26 & 497.3 & 348.0 & 149.3 \\
\emph{leonardo} & 23 & 488.5 & 384.1 & 104.4 \\
\emph{immigration} & 17 & 485.3 & 374.0 & 111.3 \\
\emph{anarchy} & 12 & 484.7 & 391.3 & 93.4 \\
\emph{retail} & 10 & 482.1 & 344.5 & 137.6 \\
\emph{wisconsin} & 10 & 488.7 & 401.6 & 87.1 \\
\bottomrule
\end{tabular}
\caption{
Values of $n_t$, $Q(X_t)$, $M(X_t)$, and $V(X_t)$ for the tokens in Fig.~\ref{fig:intro}. 
See Appendix~\ref{app:intro} for details on the token selection method.
}
\label{tab:selected_tokens}
\end{table}

In this section, we explain the selection process of the tokens used in Fig.~\ref{fig:intro}. 
To account for the effect of frequency, we used only tokens with $1 \leq \log_{10} n_t \leq 5$, and for readability, we limited the selection to tokens with at least 3 characters.
Based on these conditions, we defined the interval $[\min_{t} \log_{10} n_t, \max_{t} \log_{10} n_t]$ and divided it into 10 equal subintervals.
Let $I_r$ denote the $r$-th subinterval, where $r \in \{1,\ldots,10\}$. 
From each $I_r$, we defined the set of tokens as:
\begin{align*}
T_r := \{t \in T \mid \log_{10}(n_t) \in I_r \}.
\end{align*}
Let $|T_r|$ denote the number of tokens in $T_r$. 
For each $T_r$, we ramdomly sampled
\begin{align*}
N_r := 2 + \left\lfloor \sqrt{\frac{4\,|T_r|}{\max_r{|T_r|}}} \right\rfloor
\end{align*}
tokens. 
Here, $\lfloor \cdot \rfloor$ represents the floor function. 
The definition of $N_r$ is ad hoc, ensuring that at least two tokens are sampled from each $T_r$, with additional tokens sampled proportionally to $|T_r|$.

In Fig.~\ref{fig:MV_for_intro}, the selected tokens (i.e., the tokens shown in Fig.~\ref{fig:intro}) are plotted in scatter plots of $M(X_t)$ and $V(X_t)$. 

Table~\ref{tab:selected_tokens} shows the values of $n_t$, $Q(X_t)$, $M(X_t)$, and $V(X_t)$ for these tokens. 
Despite large differences in frequency, these tokens have similar values for $Q(X_t)$.

For the PCA transformation used in Fig.~\ref{fig:intro}, we also transformed the origin $\bm{0} \in \mathbb{R}^d$ to better understand the positional relationship between each distribution of embeddings and the origin. 
After the transformation, we translated all 2D points so that the transformed origin coincided with the new origin.

\section{Details of the dataset}\label{app:datasets}
\begin{figure}[t!]
    \centering
    \includegraphics[height=5cm]{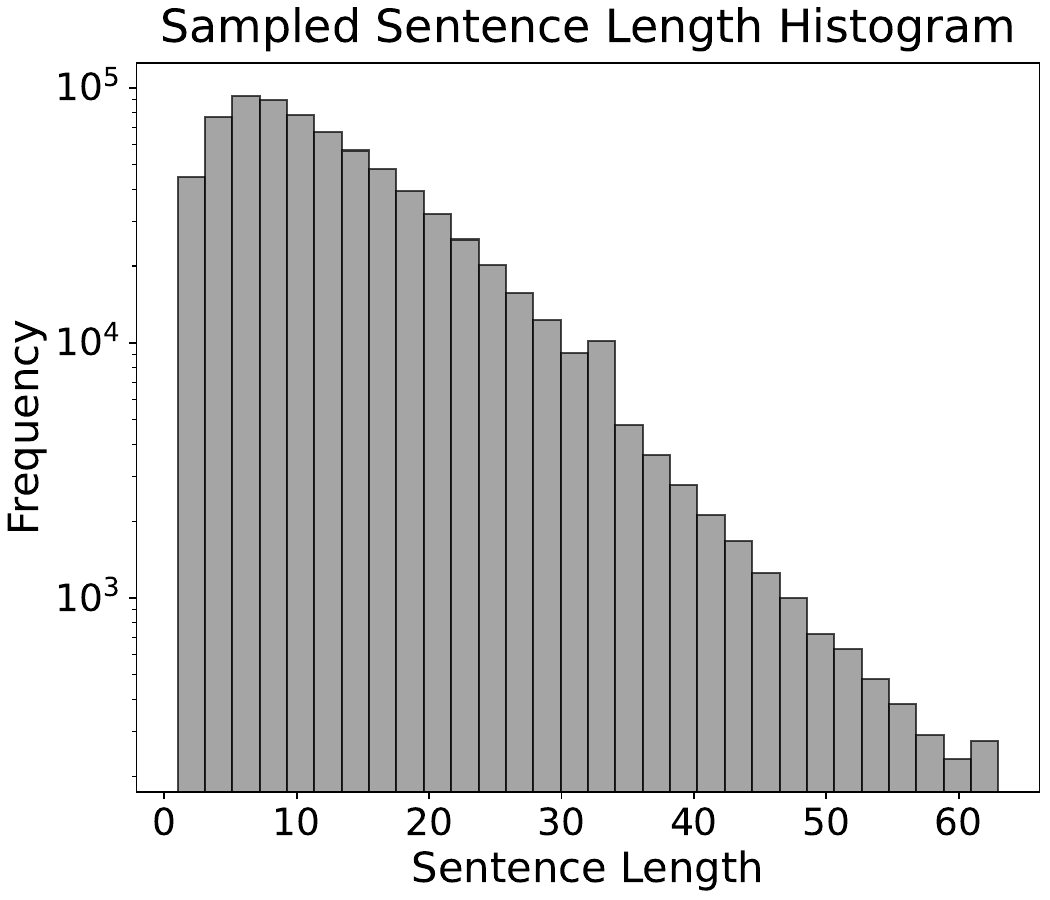}
    \caption{
Histograms of sentence lengths for the dataset we used. See Appendix~\ref{app:datasets} for details on how the sentences were selected.
}
\label{fig:sent_hist}
\end{figure}

\begin{figure}[t!]
    \centering
    \includegraphics[width=\textwidth]{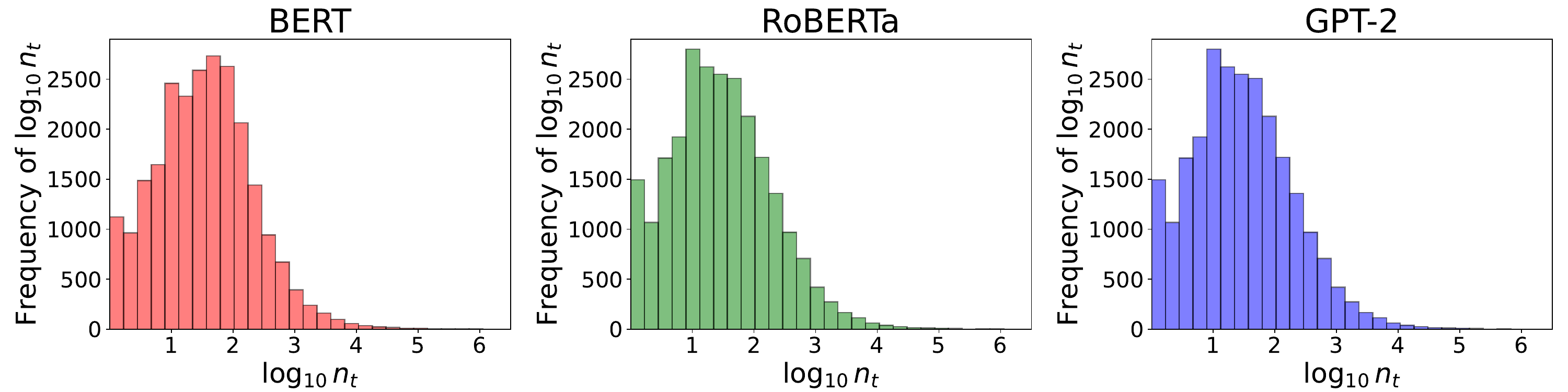}
    \caption{
Histograms of $\log_{10} n_t$ for each model. 
For differences in tokenization for each model, see Appendix~\ref{app:datasets}.
}
\label{fig:token_count}
\end{figure}

\begin{table}[t!]
\small
\centering
\begin{tabular}{lrr}
\toprule
Model & $|T|$ & $|X|$ \\
\midrule
BERT & $24{,}149$ & $12{,}384{,}011$\\
RoBERTa & $24{,}719$ & $12{,}238{,}028$\\ 
GPT-2 & $24{,}718$ & $12{,}238{,}028$\\ 
\bottomrule
\end{tabular}
\caption{
Values of $|T|$ and $|X|$ by tokenization for each model. While RoBERTa uses the same tokenizer as GPT-2, it distinguishes between the beginning-of-sentence and end-of-sentence tokens, unlike GPT-2.
}
\label{tab:T_N}
\end{table}

As described in Section~\ref{sec:datasets}, we randomly sampled 1\% of the sentences from the BookCorpus~\cite{DBLP:conf/iccv/ZhuKZSUTF15} and used sentences with fewer than 64 words for embedding calculations. The histogram of sentence lengths for the sampled sentences is shown in Fig.~\ref{fig:sent_hist}.

The tokenization of BERT differs from that of RoBERTa and GPT-2. RoBERTa uses the same tokenizer as GPT-2 but differs in that it distinguishes between beginning-of-sentence (BOS) and end-of-sentence (EOS) tokens. RoBERTa uses \emph{<s>} for BOS and \emph{</s>} for EOS, while GPT-2 uses \emph{<|endoftext|>} for both BOS and EOS. 
Table~\ref{tab:T_N} shows $|T|$ and $|X|$.
The histograms of log-scale token frequencies, $\log_{10} n_t$, for each model are shown in Fig.~\ref{fig:token_count}.

\section{Details of the statistical measures for an embedding set \texorpdfstring{$X_t$}{Xt} in Section~\ref{sec:Xt}}\label{app:Xt}
In this section, as discussed in Section~\ref{sec:Xt}, we explain the values $Q(X_t)$, $M(X_t)$, and $V(X_t)$ for an embedding set \texorpdfstring{$X_t$}{Xt}, as well as the relationship between them, given by (\ref{eq:QMVXt}). 
First, we explain the statistical measures for each component, and then we provide the proof of (\ref{eq:QMVXt}).

\subsection{Values for each component}\label{app:onedimdef}
Using the $i$-th component $x_i$ of an embedding $\bm{x}\in\mathbb{R}^d$, we define the following values for $X_t$:
\begin{align}
    q_i (X_t) &:= \mathbb{E}_{\bm{x}\in X_t} \left\{ x_i^2\right\}, \label{eq:qXt} \\
    \mu_i(X_t) &:= \mathbb{E}_{\bm{x}\in X_t} \left\{ x_i\right\}, \label{eq:muXt}\\
    v_i (X_t) &:= \mathbb{E}_{\bm{x}\in X_t} \left\{ (x_i - \mu_i (X_t) )^2\right\}, \label{eq:vXt}
\end{align}
where $\mu_i(X_t)$ is the $i$-th component of $\bm{\mu}(X_t)$ in (\ref{eq:MuXt}), and $v_i(X_t)$ is the variance of the $i$-th component of the embeddings in $X_t$, $\{x_i\mid \bm{x}\in X_t\}$. 
Then, the following relationship holds between $q_i(X_t)$, $\mu_i(X_t)$, and $v_i(X_t)$:
\begin{align}
    q_i(X_t) = \mu_i(X_t)^2 + v_i(X_t).\label{eq:qmuv}
\end{align}
\begin{proof}
\begin{align*}
v_i (X_t)
&= \mathbb{E}_{\bm{x}\in X_t} \left\{ (x_i - \mu_i (X_t) )^2\right\} \\
&= \mathbb{E}_{\bm{x}\in X_t} \left\{ x_i^2 - 2\mu_i(X_t) x_i + \mu_i(X_t)^2 \right\} \\
&= \mathbb{E}_{\bm{x}\in X_t} \left\{ x_i^2 \right\} - \mu_i(X_t)^2\\
&= q_i (X_t) - \mu_i (X_t)^2.
\end{align*}
\end{proof}
This is nothing more than the well-known formula for variance in elementary statistics.

Thus, $Q(X_t)$, $M(X_t)$, and $V(X_t)$ are the sums of $q_i(X_t)$, $\mu_i(X_t)^2$, and $v_i(X_t)$ across all components, as follows:
\begin{align}
    Q(X_t) &= \sum_{i=1}^d q_i (X_t), \label{eq:QFsum}\\
    M(X_t) &= \sum_{i=1}^d \mu_i (X_t)^2, \label{eq:MFsum}\\
    V(X_t) &= \sum_{i=1}^d v_i (X_t).\label{eq:VFsum}
\end{align}
\begin{proof}
From the definition of the $L_2$ norm:
\begin{align*}
\|\bm{x}\|^2 = \sum_{i=1}^d x_i^2.
\end{align*}
Then we obtain:
\begin{align*}
Q(X_t)
&= \mathbb{E}_{\bm{x}\in X_t} \left\{ \|\bm{x}\|^2\right\}\\
&= \mathbb{E}_{\bm{x}\in X_t} \left\{ \sum_{i=1}^d x_i^2\right\}\\
&= \sum_{i=1}^d \mathbb{E}_{\bm{x}\in X_t} \left\{ x_i^2\right\}\\
&= \sum_{i=1}^d q_i (X_t),\\
M(X_t)
&= \|\mathbb{E}_{\bm{x}\in X_t} \left\{ \bm{x}\right\}\|^2\\
&= \|\bm{\mu}(X_t)\|^2\\
&= \sum_{i=1}^d \mu_i(X_t)^2,\\
V(X_t)
&= \mathbb{E}_{\bm{x}\in X_t} \left\{ \|\bm{x} - \bm{\mu} (X_t) \|^2\right\}\\
&= \mathbb{E}_{\bm{x}\in X_t} \left\{ \sum_{i=1}^d \left(x_i-\mu_i(X_t)\right)^2 \right\}\\
&= \sum_{i=1}^d \mathbb{E}_{\bm{x}\in X_t} \left\{ \left(x_i-\mu_i(X_t)\right)^2 \right\}\\
&= \sum_{i=1}^d v_i(X_t).\\
\end{align*}
\end{proof}

\subsection{Proof of (\ref{eq:QMVXt}) related to \texorpdfstring{$Q(X_t)$}{QXt}, \texorpdfstring{$M(X_t)$}{MXt}, and \texorpdfstring{$V(X_t)$}{VXt}}\label{app:QMVXt}
We prove (\ref{eq:QMVXt}) given by:
\begin{align*}
 Q(X_t) = M(X_t) + V(X_t).
\end{align*}
\begin{proof}
By summing both sides of (\ref{eq:qmuv}) from $i = 1$ to $d$:
\begin{align*}
\sum_{i=1}^d q_i(X_t) &= \sum_{i=1}^d \mu_i(X_t)^2 + \sum_{i=1}^d v_i(X_t).
\end{align*}
Then we obtain the result using (\ref{eq:QFsum}), (\ref{eq:MFsum}), and (\ref{eq:VFsum}). 
Alternatively, the result can be derived directly as follows:
\begin{align*}
V(X_t)
&= \mathbb{E}_{\bm{x}\in X_t} \left\{ \|\bm{x} - \bm{\mu} (X_t) \|^2\right\}\\
&= \mathbb{E}_{\bm{x}\in X_t} \left\{(\bm{x} - \bm{\mu} (X_t))^\top (\bm{x} - \bm{\mu} (X_t)) \right\}\\
&= \mathbb{E}_{\bm{x}\in X_t} \left\{\|\bm{x}\|^2 - \bm{x}^\top\bm{\mu} (X_t) - \bm{\mu} (X_t)^\top\bm{x} + \|\bm{\mu} (X_t)\|^2 \right\}\\
&= \mathbb{E}_{\bm{x}\in X_t} \left\{\|\bm{x}\|^2  \right\} - \mathbb{E}_{\bm{x}\in X_t} \left\{ \bm{x}\right\}^\top\bm{\mu} (X_t)  - \bm{\mu} (X_t)^\top\mathbb{E}_{\bm{x}\in X_t} \left\{  \bm{x}\right\} + \|\bm{\mu} (X_t)\|^2\\
&= \mathbb{E}_{\bm{x}\in X_t} \left\{\|\bm{x}\|^2  \right\} - \|\bm{\mu} (X_t)\|^2\\
&= Q(X_t) - M(X_t).
\end{align*}     
\end{proof}

\section{Proof of (\ref{eq:VwVb}) decomposing \texorpdfstring{$V(X)$}{VX} into \texorpdfstring{$V_W(X)$}{VWX} and \texorpdfstring{$V_B(X)$}{VBX}}\label{app:VwVb}

In this section, we will prove (\ref{eq:VwVb}). 
To do so, we first prove the following equation for $v_i(X)$, where $v_i(X)$ is the value obtained by replacing $X_t$ with $X$ in $v_i(X_t)$ as defined in (\ref{eq:vXt}) in Appendix~\ref{app:Xt}:
\begin{align}
v_i(X)=\sum_{t\in T} p_t \left\{v_i(X_t) + (\mu_i(X_t)-\mu_i(X))^2\right\}.\label{eq:vwvb}
\end{align}
\begin{proof}
\begin{align*}
v_i(X)
&= \mathbb{E}_{\bm{x}\in X} \left\{ (x_i - \mu_i (X) )^2\right\} \\
&= \frac{1}{|X|}\sum_{\bm{x}\in X} (x_i - \mu_i (X) )^2\\
&= \frac{1}{|X|}\sum_{t\in T}\sum_{\bm{x}\in X_t} (x_i - \mu_i (X) )^2\\
&= \sum_{t\in T} \frac{|X_t|}{|X|}\cdot\frac{1}{|X_t|} \sum_{\bm{x}\in X_t} (x_i - \mu_i (X) )^2\\
&= \sum_{t\in T} p_t \mathbb{E}_{\bm{x}\in X_t} \left\{ (x_i - \mu_i(X_t) + \mu_i(X_t) - \mu_i (X) )^2\right\} \\
&= \sum_{t\in T} p_t \mathbb{E}_{\bm{x}\in X_t} \left\{ (x_i - \mu_i(X_t))^2 - 2 (x_i - \mu_i(X_t))(\mu_i(X_t) - \mu_i (X) )  + (\mu_i(X_t) - \mu_i (X) )^2\right\} \\
&= \sum_{t\in T} p_t \left\{\mathbb{E}_{\bm{x}\in X_t} \left\{ (x_i - \mu_i(X_t))^2\right\} + (\mu_i(X_t) - \mu_i (X) )^2 \right\} \quad(\because \mu_i(X_t) = \mathbb{E}_{\bm{x}\in X_t} \left\{ x_i\right\})\\
&= \sum_{t\in T} p_t \left\{v_i(X_t) + (\mu_i(X_t) - \mu_i (X) )^2 \right\}.
\end{align*}
\end{proof}

Based on (\ref{eq:vwvb}), we prove (\ref{eq:VwVb}) given by:
\begin{align*}
V(X) = V_W(X) + V_B(X).
\end{align*}
\begin{proof}
\begin{align*}
V(X)
&=\sum_{i=1}^d v_i(X) \quad\text{(from (\ref{eq:VFsum}), where $X_t$ is replaced by $X$)}\\
&=\sum_{i=1}^d \sum_{t\in T} p_t \left\{v_i(X_t) + (\mu_i(X_t)-\mu_i(X))^2 \right\}\\
&=\sum_{t\in T} p_t \sum_{i=1}^d v_i(X_t) + \sum_{t\in T} p_t \sum_{i=1}^d (\mu_i(X_t)-\mu_i(X))^2 \\
&=\sum_{t\in T} p_t V(X_t) + \sum_{t\in T} p_t \|\bm{\mu}(X_t) - \bm{\mu}(X)\|^2\quad(\because(\ref{eq:VFsum}))\\
&= V_W(X) + V_B(X).
\end{align*}
\end{proof}

\section{Details of the statistical measures for the entire embedding set \texorpdfstring{$X$}{X} in Section~\ref{sec:calcX}}\label{app:calcX}
In this section, as discussed in Section~\ref{sec:calcX}, we explain the values $n$, $\bm{\mu}(X)$, and $Q(X)$ for the entire embedding set $X$.

\subsection{Calculation of \texorpdfstring{$n$}{n}}
We define $n=|X|$ as the total number of embeddings in $X$.
We assume that $X_t$ and $X_{t'}$ are disjoint for $t, t' \in T$, i.e., $X_t\cap X_{t'}=\emptyset$. 
Then
\begin{align*}
n = \left|\bigcup_{t\in T}X_t \right| =\sum_{t\in T}|X_t|=\sum_{t\in T} n_t.
\end{align*}
Thus, $n$ can be expressed using $n_t$.

\subsection{Calculation of \texorpdfstring{$\bm{\mu}(X)$}{muX}}\label{app:MuX}
By replacing $X_t$ with $X$ in $\bm{\mu}(X_t)$ in (\ref{eq:MuXt}), the mean embedding $\bm{\mu}(X) \in \mathbb{R}^d$ for $X$ is defined. 
We prove (\ref{eq:MuX}) given by:
\begin{align*}
\bm{\mu}(X)=\mathbb{E}_{\bm{x}\in X} \left\{\bm{x}\right\}=\sum_{t\in T} p_t\bm{\mu}(X_t).
\end{align*}
\begin{proof}
\begin{align*}
\bm{\mu}(X)
&= \mathbb{E}_{\bm{x}\in X}\{\bm{x}\}\\
&= \frac{1}{|X|}\sum_{\bm{x}\in X} \bm{x}\\
&= \frac{1}{|X|}\sum_{t\in T}\sum_{\bm{x}\in X_t}\bm{x}\\
&= \sum_{t\in T} \frac{|X_t|}{|X|}\cdot\frac{1}{|X_t|}\sum_{\bm{x}\in X_t}\bm{x}\\
&= \sum_{t\in T} p_t\mathbb{E}_{\bm{x}\in X_t}\{\bm{x}\}\\
&= \sum_{t\in T} p_t\bm{\mu}(X_t).
\end{align*}
\end{proof}

\subsection{Calculation of \texorpdfstring{$Q(X)$}{QX}}\label{app:QMVX}
By replacing $X_t$ with $X$ in $Q(X_t)$ in (\ref{eq:QXt}), $Q(X)$ for $X$ is defined. 
We prove (\ref{eq:QX}) given by:
\begin{align*}
    Q(X) = \mathbb{E}_{\bm{x}\in X} \left\{ \|\bm{x}\|^2\right\} = \sum_{t\in T} p_t Q(X_t).
\end{align*}
To do so, we first prove the following equation for $q_i(X)$, where $q_i(X)$ is the value obtained by replacing $X_t$ with $X$ in $q_i(X_t)$ as defined in (\ref{eq:qXt}) in Appendix~\ref{app:Xt}:
\begin{align}
q_i(X) =\sum_{t\in T} p_t q_i(X_t). \label{eq:qX}
\end{align}
\begin{proof}
\begin{align*}
q_i(X)
&=\mathbb{E}_{\bm{x}\in X}\{x_i^2\}\\
&=\frac{1}{|X|}\sum_{\bm{x}\in X} x_i^2\\
&=\frac{1}{|X|}\sum_{t\in T}\sum_{\bm{x}\in X_t} x_i^2\\
&=\sum_{t\in T} \frac{|X_t|}{|X|}\cdot\frac{1}{|X_t|}\sum_{\bm{x}\in X_t}x_i^2\\
&=\sum_{t\in T} p_t \mathbb{E}_{\bm{x}\in X_t}\{x_i^2\}\\
&=\sum_{t\in T} p_t q_i(X_t).\\
\end{align*}
\end{proof}

Using (\ref{eq:qX}), we then show (\ref{eq:QX}):
\begin{align*}
Q(X)
&=\sum_{i=1}^d q_i(X) \quad\text{(from (\ref{eq:QFsum}), where $X_t$ is replaced by $X$)}\\
&=\sum_{i=1}^d \sum_{t\in T} p_t q_i(X_t)\quad(\because(\ref{eq:qX}))\\
&=\sum_{t\in T} p_t \sum_{i=1}^d q_i(X_t)\\
&=\sum_{t\in T} p_t Q(X_t)\quad(\because(\ref{eq:QFsum})).
\end{align*}

\section{Relationships between frequency and \texorpdfstring{$Q(X_t)$}{QXt}, \texorpdfstring{$M(X_t)$}{MXt}, and \texorpdfstring{$V(X_t)$}{VXt}}\label{app:QMV_MV}
\begin{figure*}[p]
\centering
\begin{subfigure}{\textwidth}
\centering
    \includegraphics[height=9.5cm, keepaspectratio]{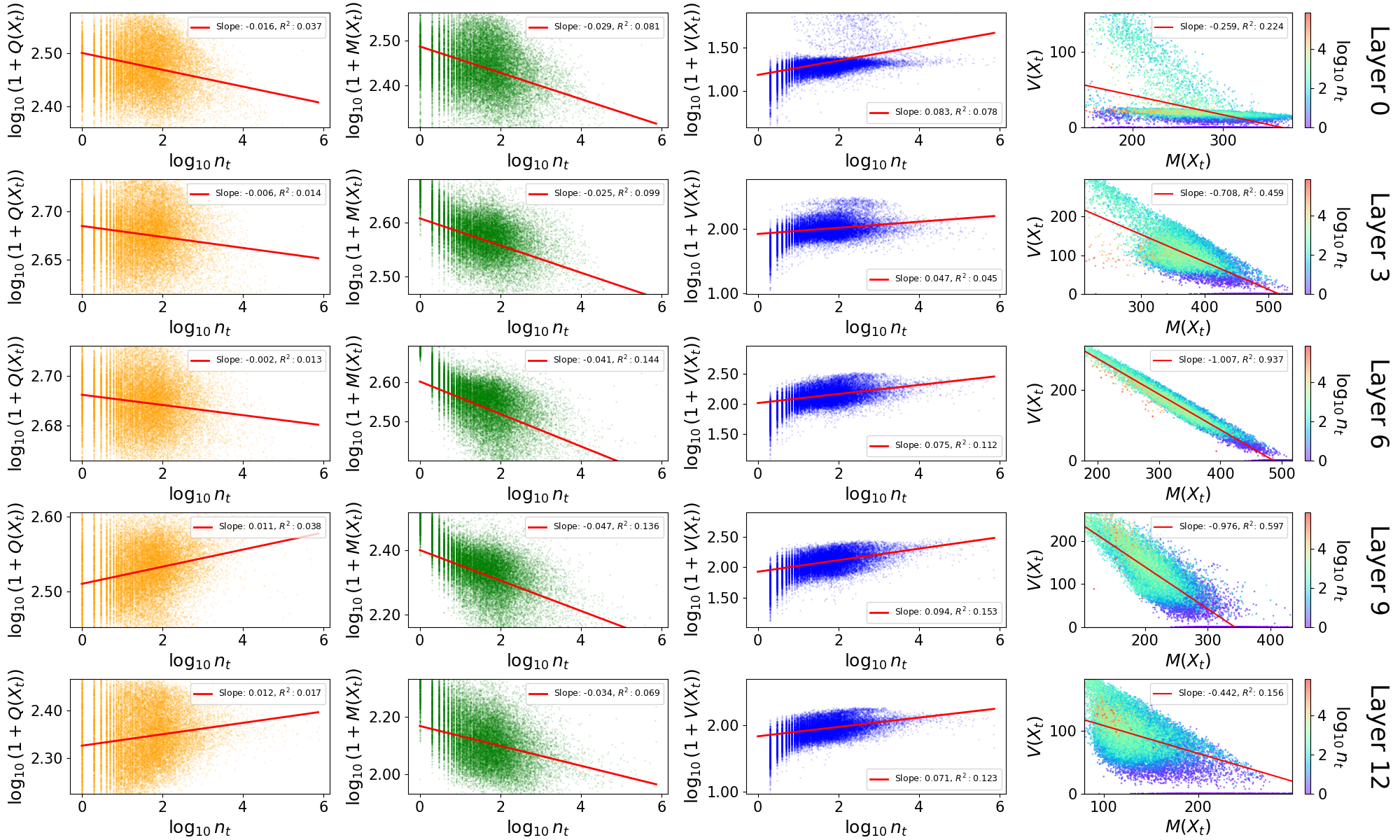}
    \caption{bert-base-uncased}
    \label{fig:QMV_MV_bert_base}
\end{subfigure}
\par\bigskip
\begin{subfigure}{\textwidth}
\centering
    \includegraphics[height=9.5cm, keepaspectratio]{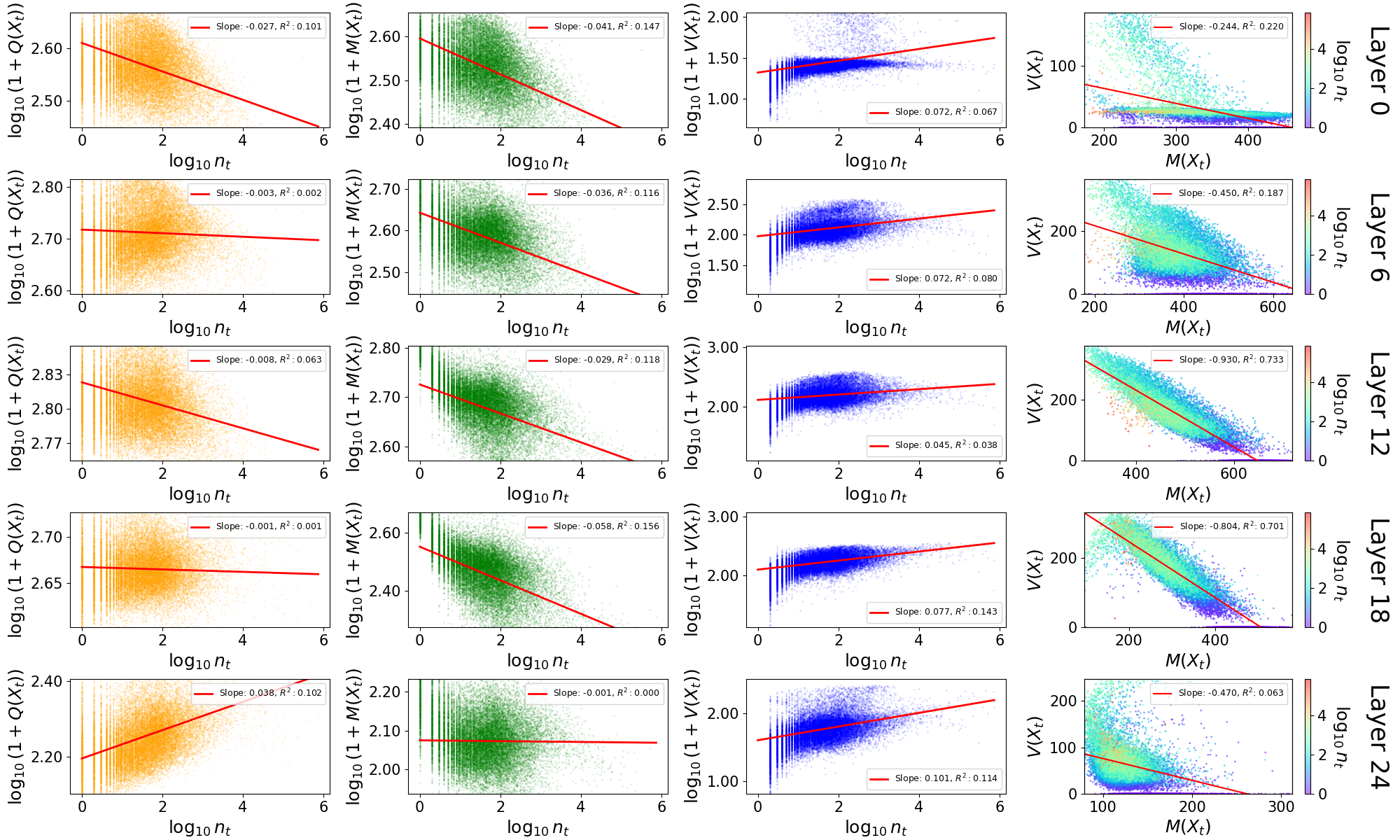}
    \caption{bert-large-uncased}
    \label{fig:QMV_MV_bert_large}
\end{subfigure}
\caption{
Scatter plots across some layers from the input to the output layer of (a) \texttt{bert-base-uncased} and (b)~\texttt{bert-large-uncased}, plotting $\log_{10}(1+Q(X_t))$, $\log_{10}(1+M(X_t))$, and $\log_{10}(1+V(X_t))$ against $\log_{10} n_t$, and plotting $V(X_t)$ against $M(X_t)$.
Each plot includes a regression line, its slope, and the coefficient of determination, $R^2$.
Only tokens with $1 \leq \log_{10} n_t \leq 5$ were used for regressions to reduce the influence of extreme values.
}
\label{fig:QMV_MV_bert_both}
\end{figure*}

\begin{figure*}[p]
\centering
\begin{subfigure}{\textwidth}
\centering
    \includegraphics[height=9.5cm, keepaspectratio]{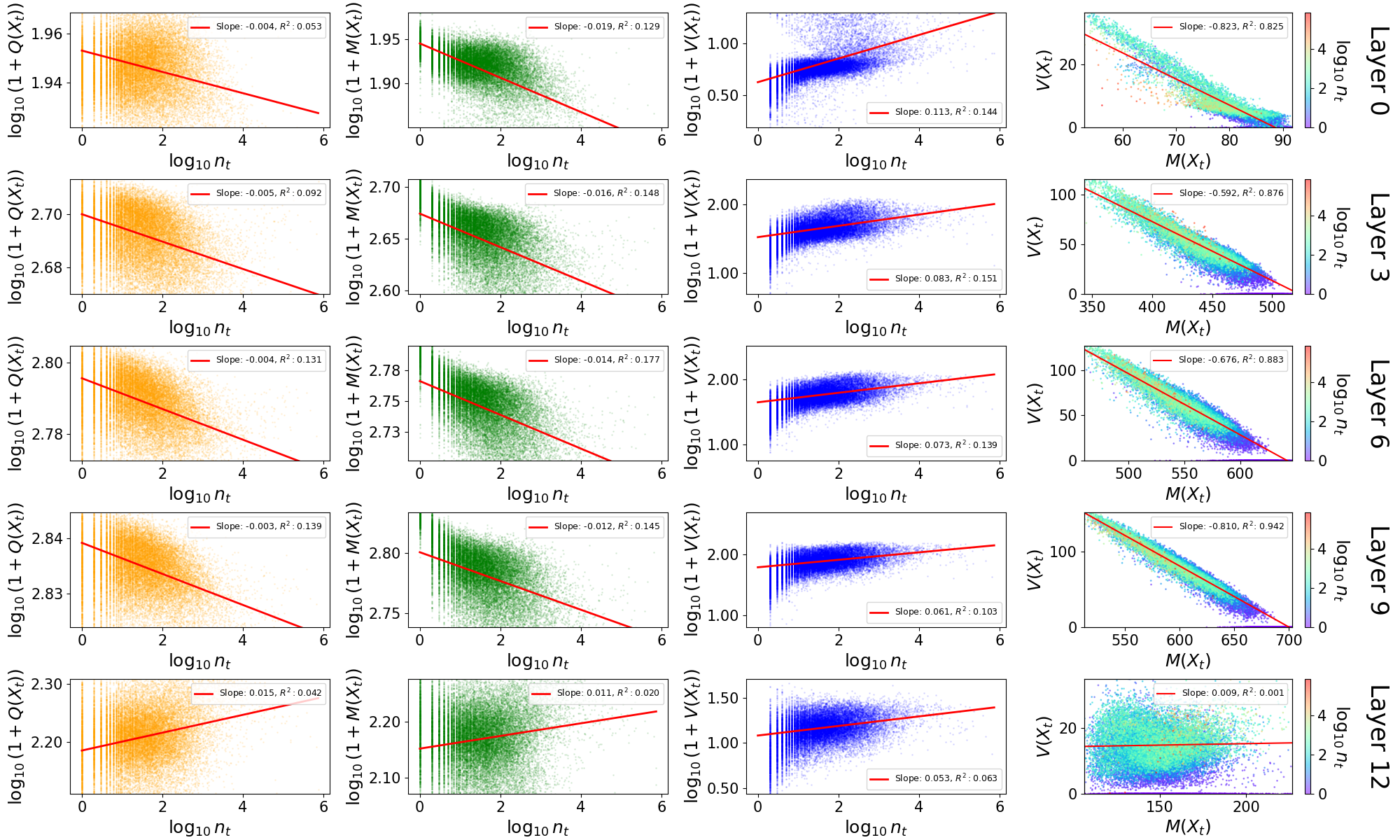}
    \caption{roberta-base}
    \label{fig:QMV_MV_roberta_base}
\end{subfigure}
\par\bigskip
\begin{subfigure}{\textwidth}
\centering
    \includegraphics[height=9.5cm, keepaspectratio]{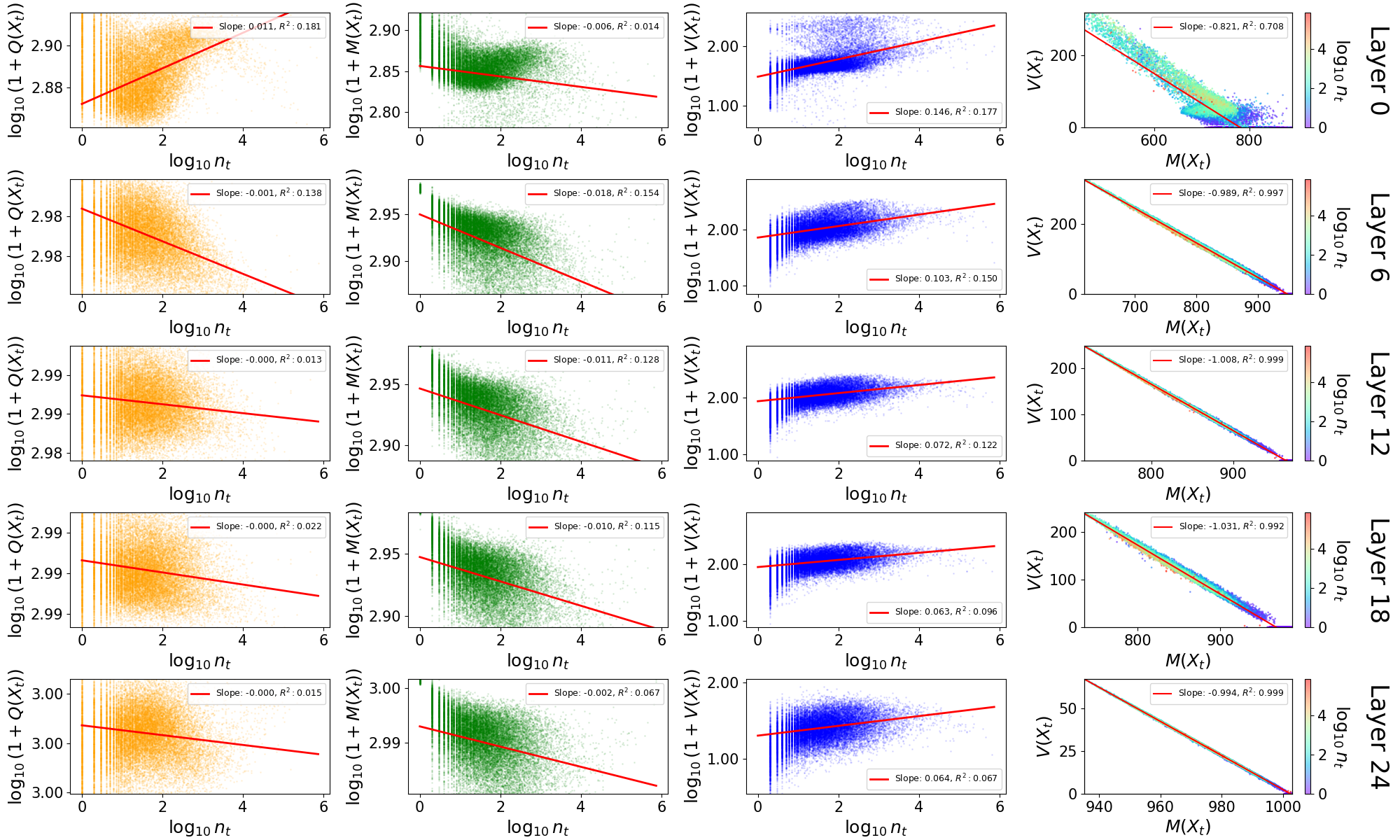}
    \caption{roberta-large}
    \label{fig:QMV_MV_roberta_large}
\end{subfigure}
\caption{
Results of the same experiments as in Fig.~\ref{fig:QMV_MV_bert_both} for (a)~\texttt{roberta-base} and (b)~\texttt{roberta-large}.
}
\label{fig:QMV_MV_roberta_both}
\end{figure*}

\begin{figure*}[p]
\centering
\begin{subfigure}{\textwidth}
\centering
    \includegraphics[height=9.5cm, keepaspectratio]{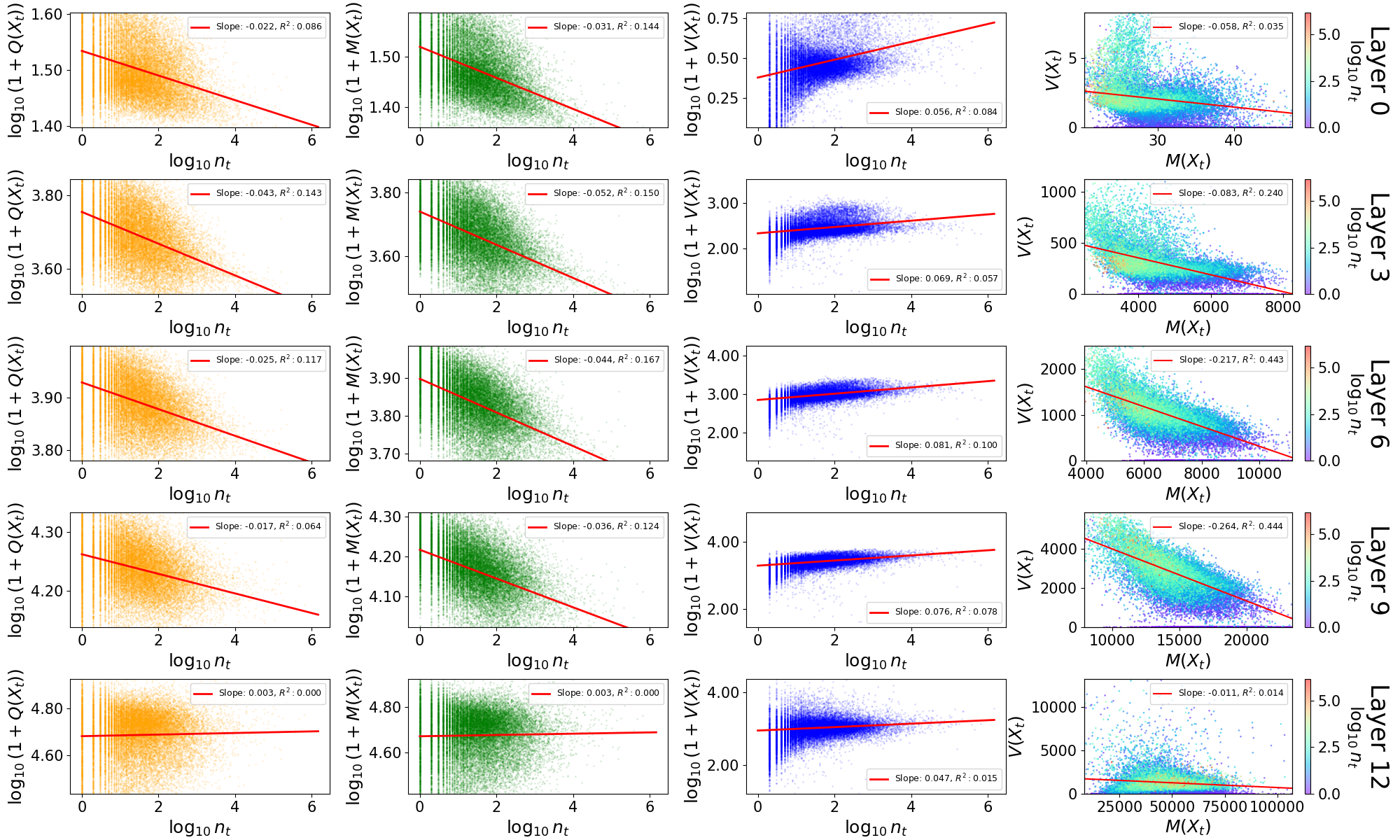}
    \caption{gpt2}
    \label{fig:QMV_MV_gpt2}
\end{subfigure}
\par\bigskip
\begin{subfigure}{\textwidth}
\centering
    \includegraphics[height=9.5cm, keepaspectratio]{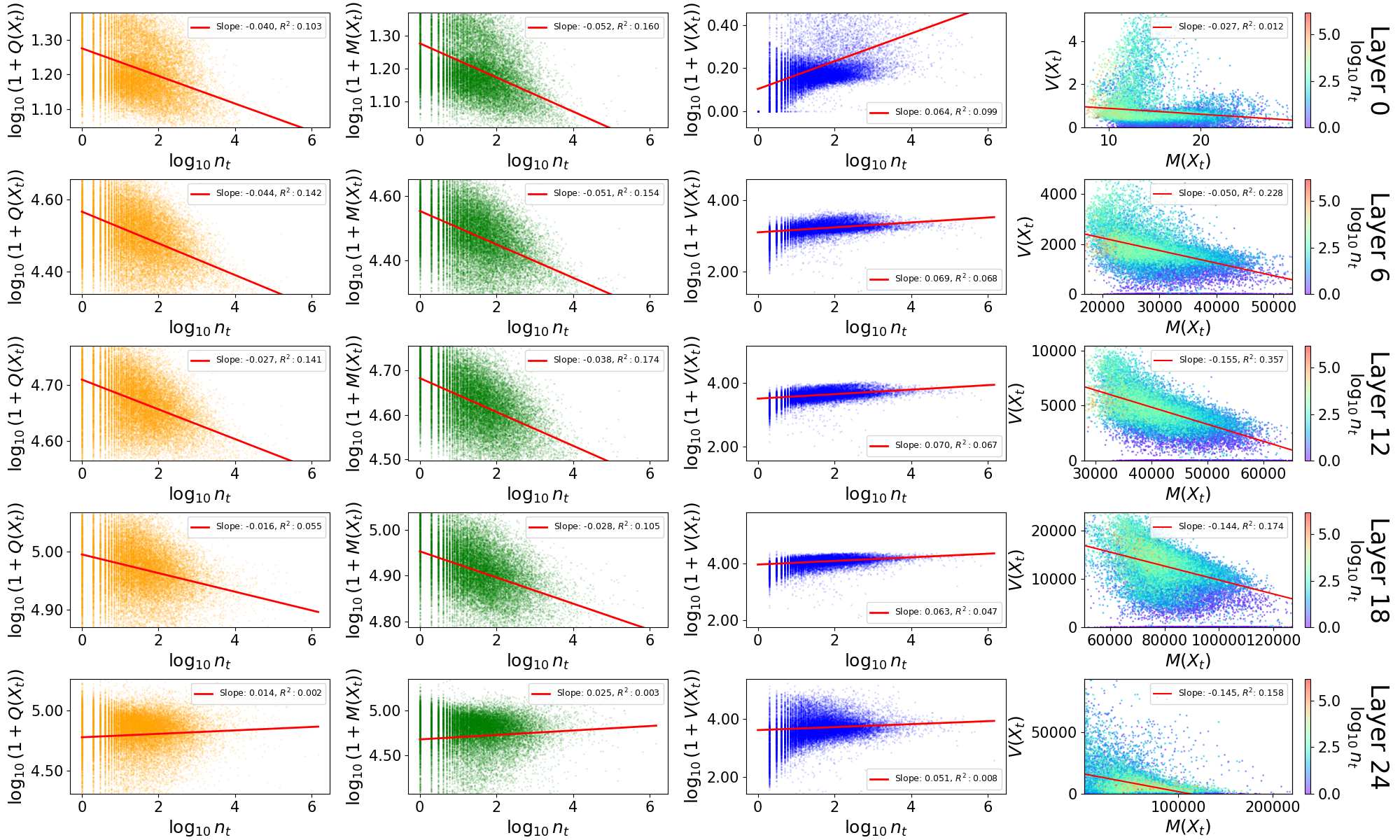}
    \caption{gpt2-medium}
    \label{fig:QMV_MV_gpt2_medium}
\end{subfigure}
\caption{
Results of the same experiments as in Fig.~\ref{fig:QMV_MV_bert_both} for (a)~\texttt{gpt2} and (b)~\texttt{gpt2-medium}.
}
\label{fig:QMV_MV_gpt2_both}
\end{figure*}

Similar to previous work~\cite{DBLP:conf/acl/Wannasuphoprasit23,DBLP:conf/icann/LiangCZRG21,DBLP:journals/corr/abs-2104-08465,DBLP:conf/acl/ZhouECJ22}, We shows scatter plots\footnote{Unlike previous work, $\log_{10}$ scales are used for $Q(X_t)$, $M(X_t)$, and $V(X_t)$ to address the large value differences.} that represent the relationships between frequency and the values of $Q(X_t)$, $M(X_t)$, and $V(X_t)$ for some layers of BERT in Fig.~\ref{fig:QMV_MV_bert_both}, RoBERTa in Fig.~\ref{fig:QMV_MV_roberta_both}, and GPT-2 in Fig.~\ref{fig:QMV_MV_gpt2_both}, respectively.
Regression lines are also shown for these plots.
The slopes of the lines for $Q(X_t)$ increase as the layers deepen, while those for $M(X_t)$ and $V(X_t)$ are consistently negative and positive, respectively. 
These trends are consistent with those observed in previous work.
Additionally, the rightmost columns of Figs.~\ref{fig:QMV_MV_bert_both},~\ref{fig:QMV_MV_roberta_both}, and~\ref{fig:QMV_MV_gpt2_both} show the scatter plots and regression lines for $M(X_t)$ and $V(X_t)$.
In the intermediate layer, the slope is smaller and the $R^2$ value is larger than that of other layers.
These results indicate a strong trade-off relationship between $M(X_t)$ and $Q(X_t)$ in the intermediate layer.

\begin{figure*}[p]
\centering
\begin{subfigure}{\textwidth}
\centering
    \includegraphics[height=9.5cm, keepaspectratio]{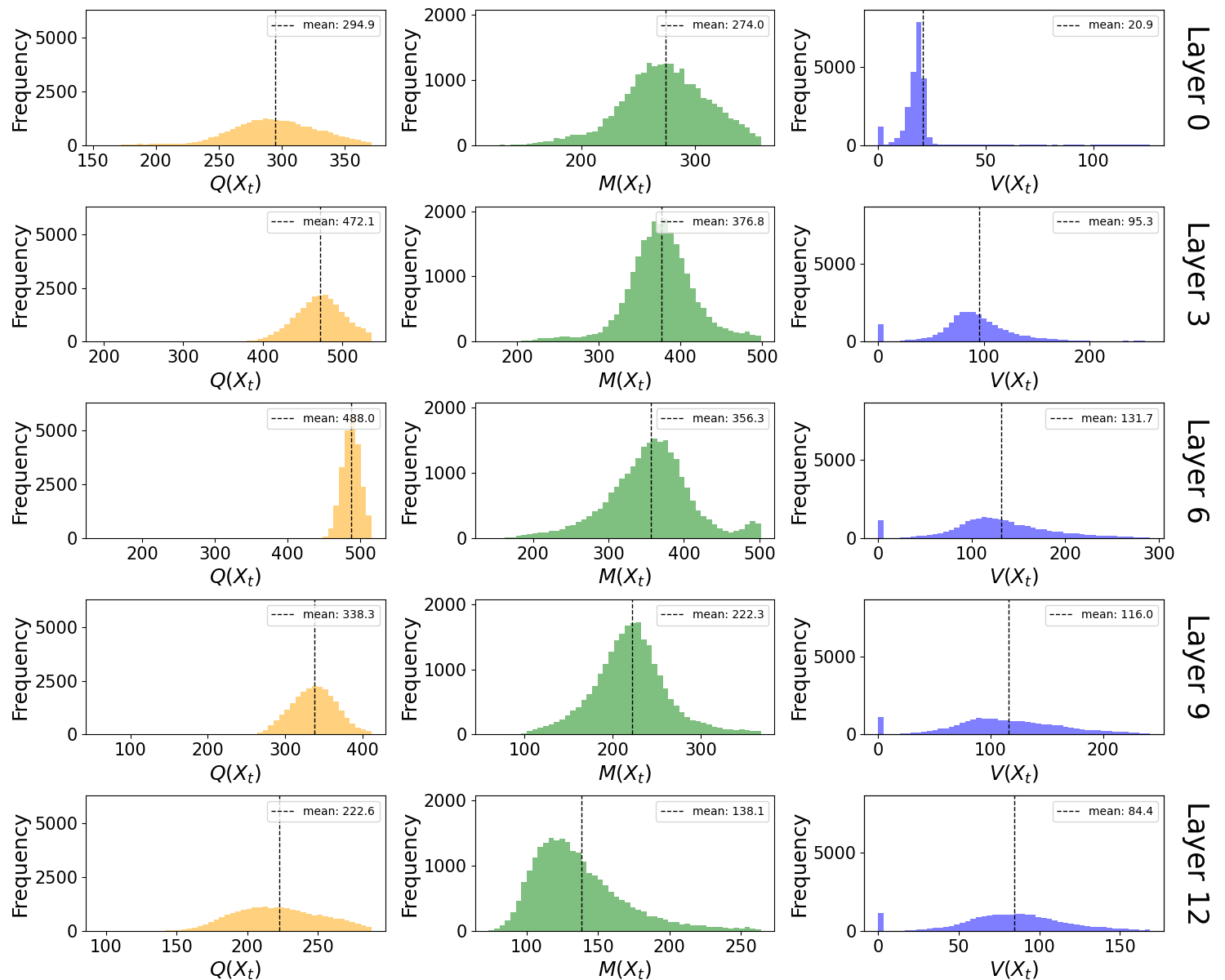}
    \caption{bert-base-uncased}
    \label{fig:QMV_hist_bert_base}
\end{subfigure}
\par\bigskip
\begin{subfigure}{\textwidth}
\centering
    \includegraphics[height=9.5cm, keepaspectratio]{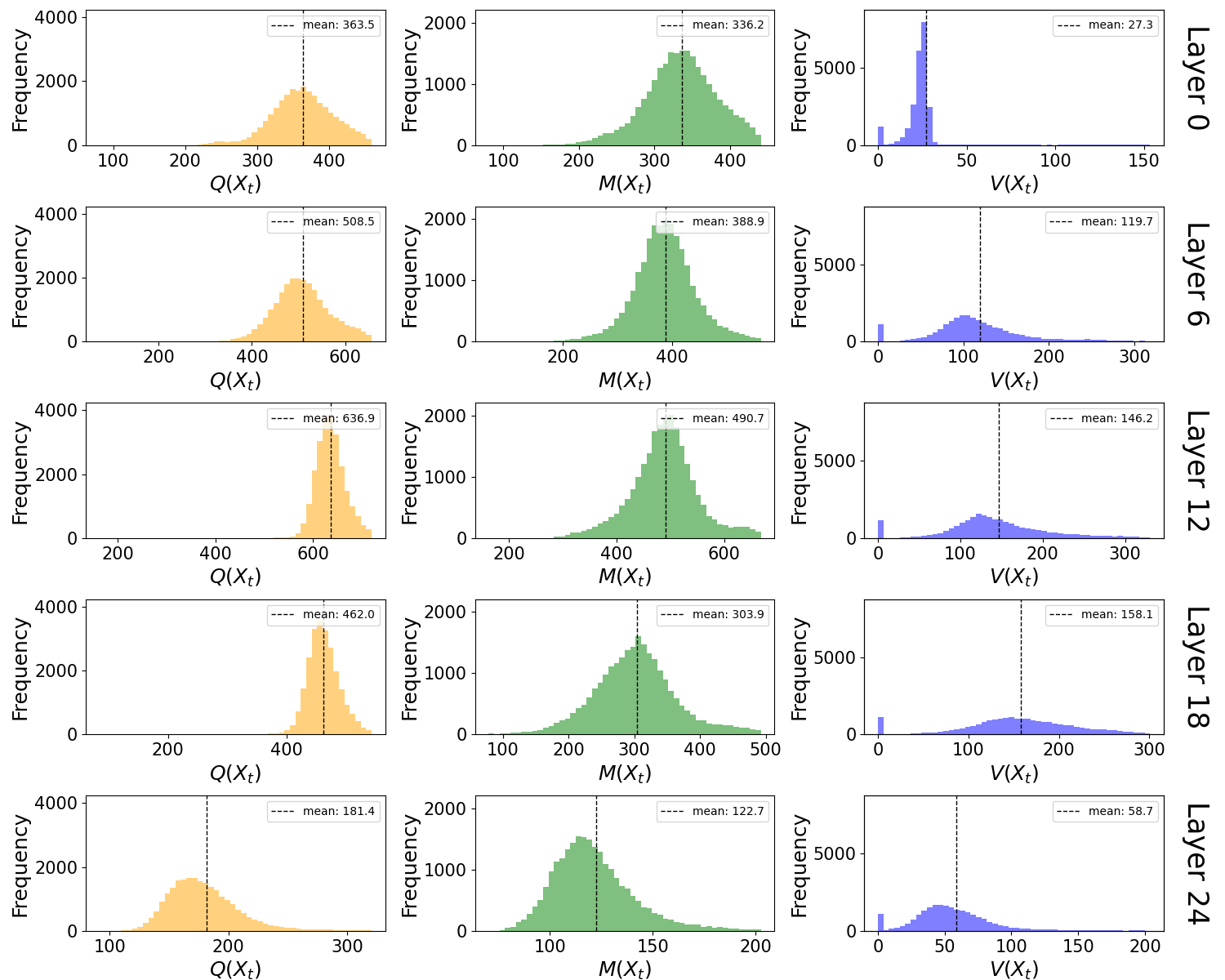}
    \caption{bert-large-uncased}
    \label{fig:QMV_hist_bert_large}
\end{subfigure}
\caption{
Histograms of $Q(X_t)$, $M(X_t)$, and $V(X_t)$ for each layer of (a)~\texttt{bert-base-uncased} and (b)~\texttt{bert-large-uncased}.
}
\label{fig:QMV_hist_bert_both}
\end{figure*}

\begin{figure*}[p]
\centering
\begin{subfigure}{\textwidth}
\centering
    \includegraphics[height=9.5cm, keepaspectratio]{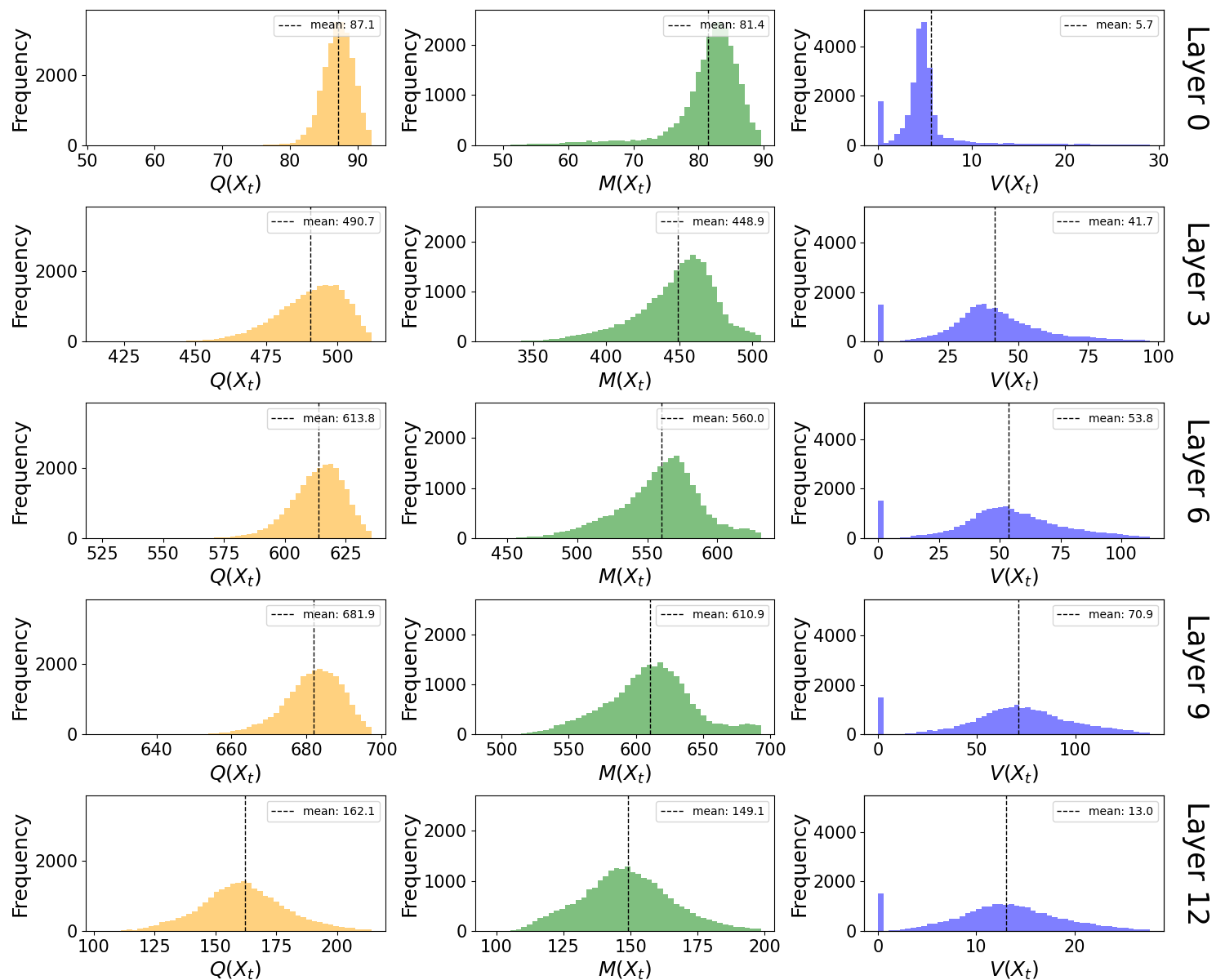}
    \caption{roberta-base}
    \label{fig:QMV_hist_roberta_base}
\end{subfigure}
\par\bigskip
\begin{subfigure}{\textwidth}
\centering
    \includegraphics[height=9.5cm, keepaspectratio]{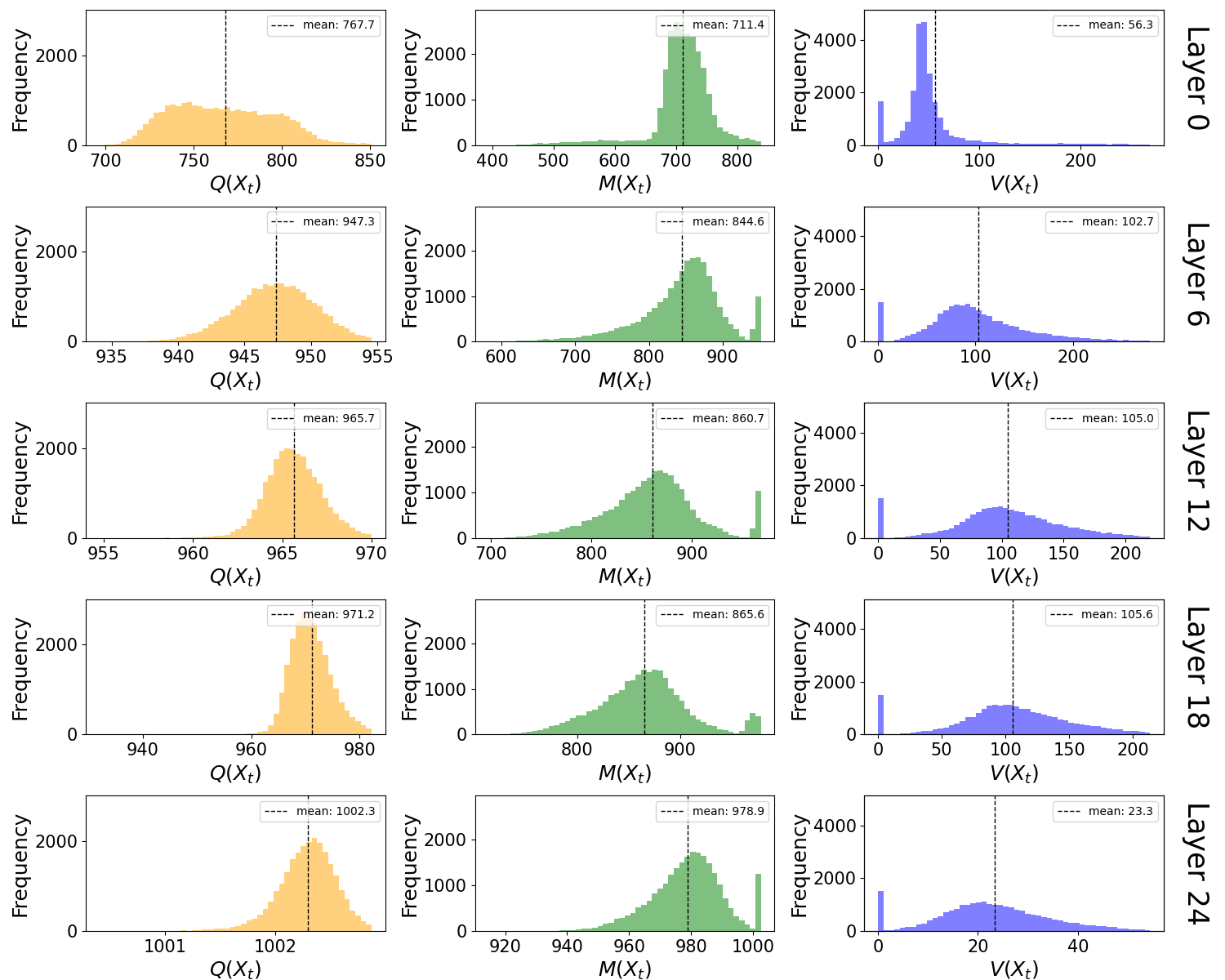}
    \caption{roberta-large}
    \label{fig:QMV_hist_roberta_large}
\end{subfigure}
\caption{
Histograms of $Q(X_t)$, $M(X_t)$, and $V(X_t)$ for each layer of (a)~\texttt{roberta-base} and (b)~\texttt{roberta-large}.
}
\label{fig:QMV_hist_roberta_both}
\end{figure*}

\begin{figure*}[p]
\centering
\begin{subfigure}{\textwidth}
\centering
    \includegraphics[height=9.5cm, keepaspectratio]{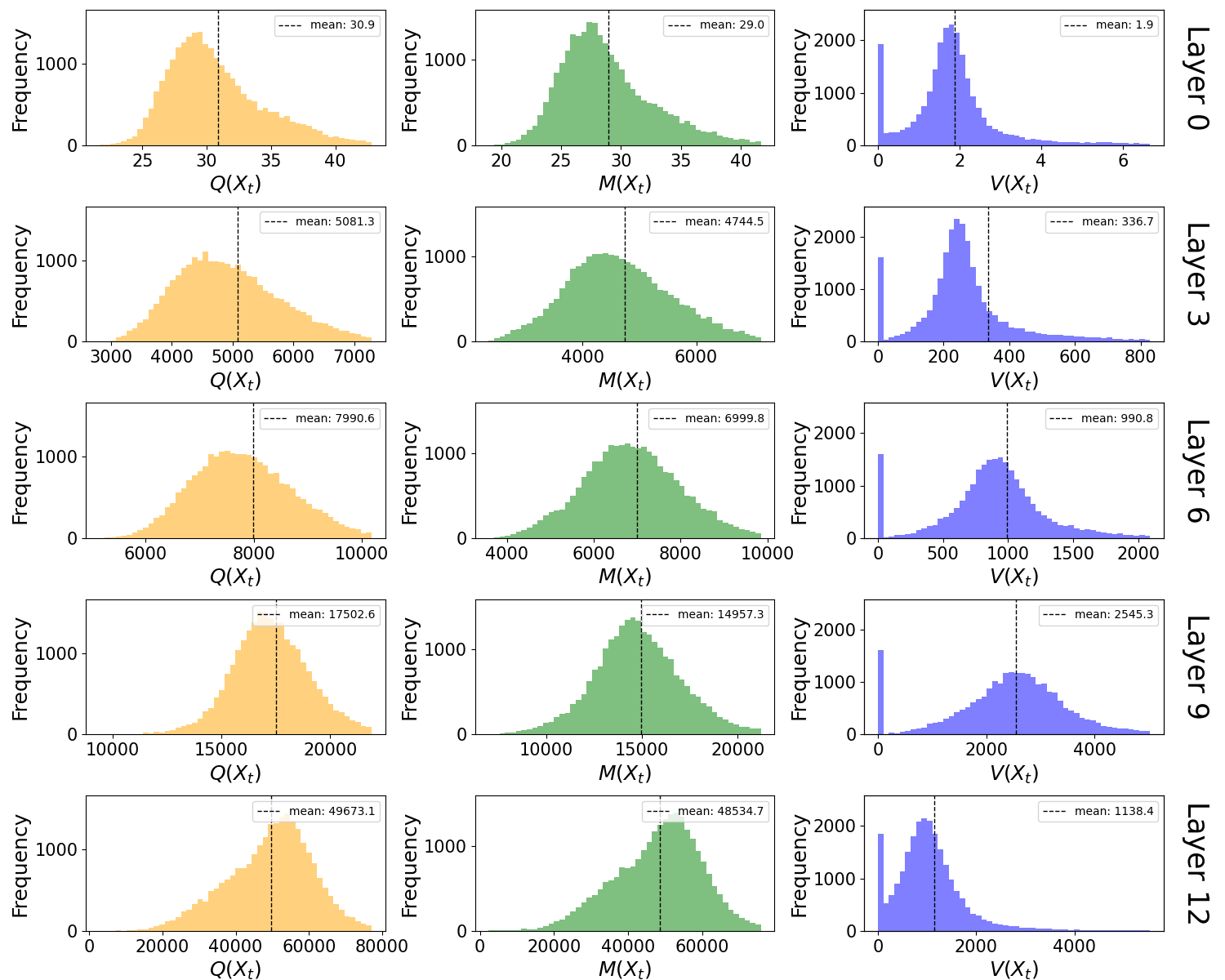}
    \caption{gpt2}
    \label{fig:QMV_hist_gpt2}
\end{subfigure}
\par\bigskip
\begin{subfigure}{\textwidth}
\centering
    \includegraphics[height=9.5cm, keepaspectratio]{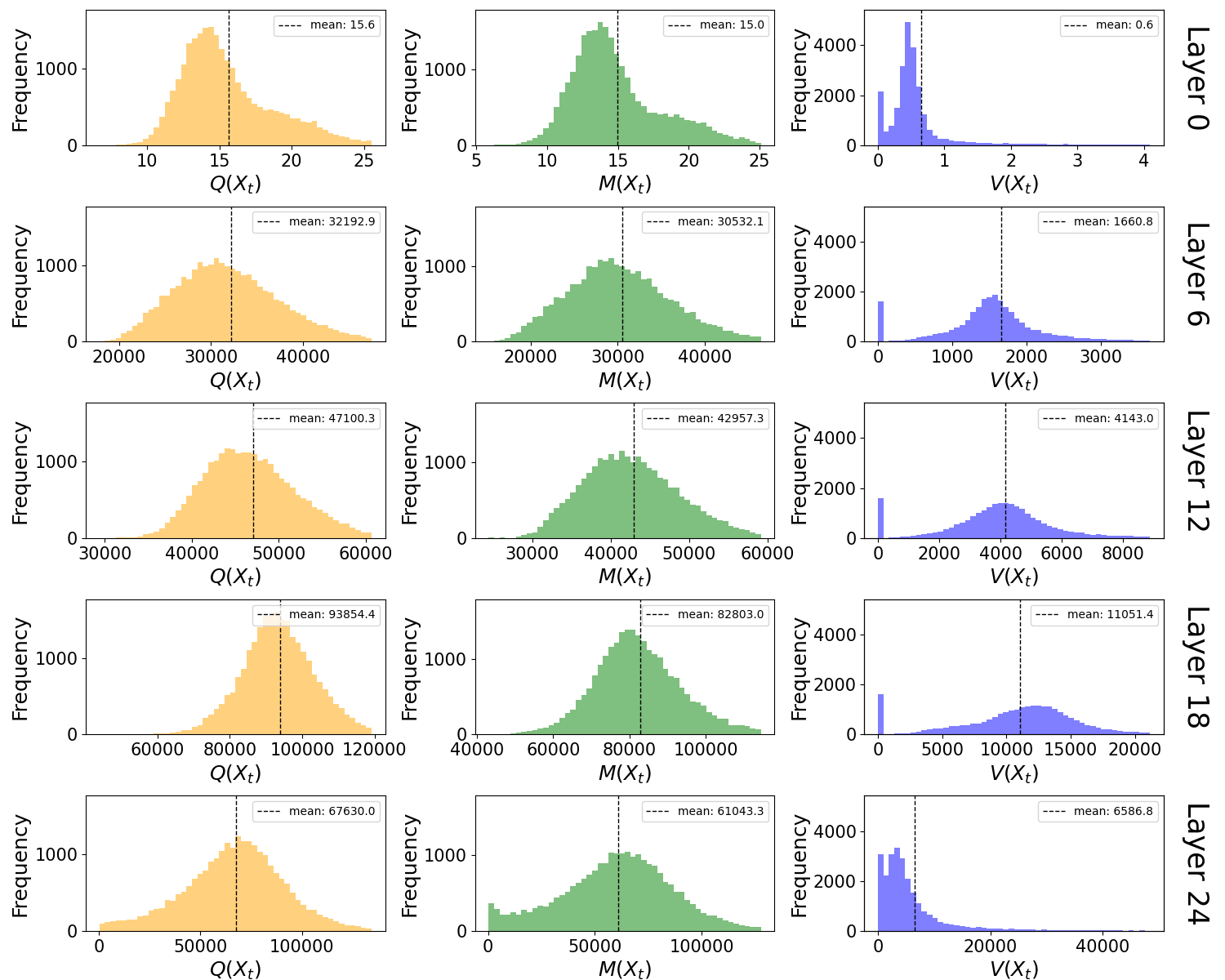}
    \caption{gpt2-medium}
    \label{fig:QMV_hist_gpt2_medium}
\end{subfigure}
\caption{
Histograms of $Q(X_t)$, $M(X_t)$, and $V(X_t)$ for each layer of (a)~\texttt{gpt2} and (b)~\texttt{gpt2-medium}.
}
\label{fig:QMV_hist_gpt2_both}
\end{figure*}

Additionally, histograms of $Q(X_t)$, $M(X_t)$, and $V(X_t)$ for the layers of the models are shown in Fig.~\ref{fig:QMV_hist_bert_both} for BERT, Fig.~\ref{fig:QMV_hist_roberta_both} for RoBERTa, and Fig.~\ref{fig:QMV_hist_gpt2_both} for GPT-2.

\begin{figure*}[p]
    \centering
    \includegraphics[width=\textwidth]{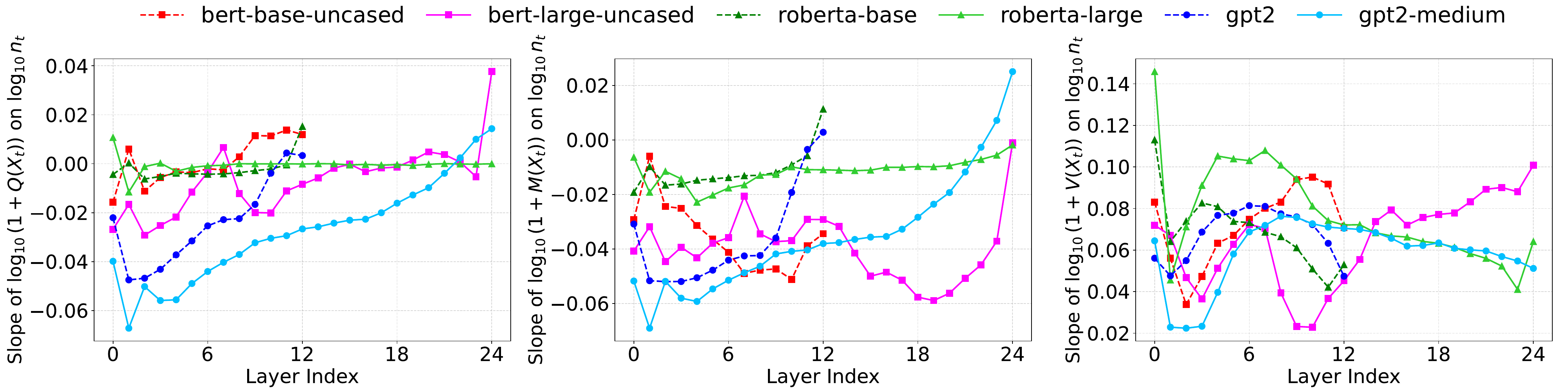}
    \caption{
Regression slopes between $\log_{10}(n_t)$ and each of $\log_{10}(1+Q(X_t))$, $\log_{10}(1+M(X_t))$, and $\log_{10}(1+V(X_t))$ across the six models and layers. 
As the layers deepen, the slopes for $Q(X_t)$ tend to increase. 
Across all layers, the slopes for $M(X_t)$ generally remain negative, while those for $V(X_t)$ remain positive. Only tokens with $1 \leq \log_{10} n_t \leq 5$ were used for regressions to reduce the influence of extreme values.
}
\label{fig:QXt_MXt_VXt_slope}
\end{figure*}

Based on Figs.~\ref{fig:QMV_MV_bert_both},~\ref{fig:QMV_MV_roberta_both}, and~\ref{fig:QMV_MV_gpt2_both}, Fig.~\ref{fig:QXt_MXt_VXt_slope} shows the slopes of the regression lines for frequency and $Q(X_t)$, $M(X_t)$, and $V(X_t)$ across the six models and layers.
The slopes for $Q(X_t)$ remain stable and close to $0$ across layers, although an increase is observed in GPT-2.
As the layers deepen, the slopes for $M(X_t)$ and $V(X_t)$ remain approximately negative and positive across all layers. 
The variation in slopes remains small and stable for all values.

\section{Detailed results of the statistical measures for \texorpdfstring{$X$}{X}}\label{app:X}
\begin{figure*}[p]
    \centering
    \includegraphics[width=\textwidth]{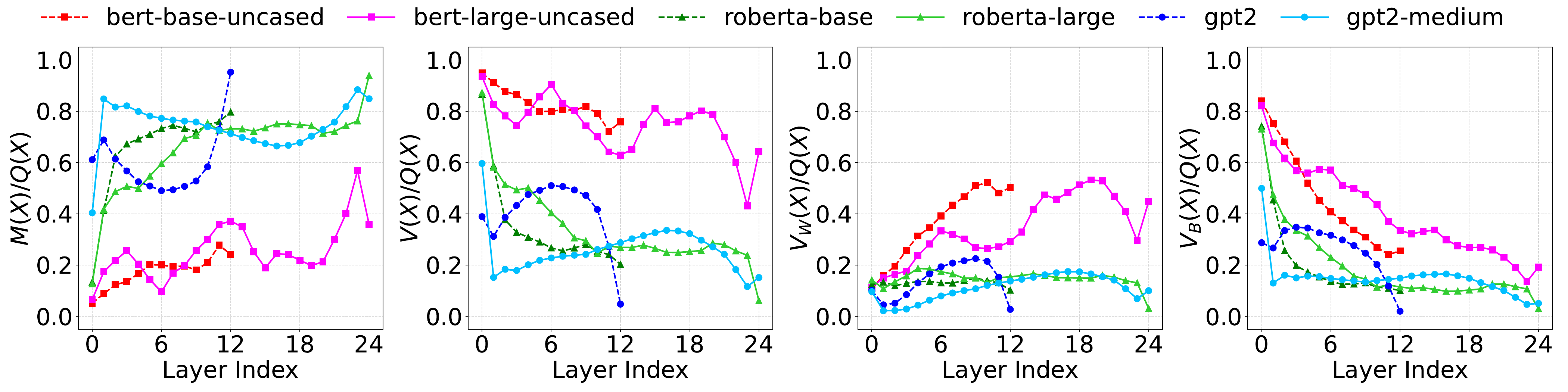}
    \caption{
The values of $M(X)$, $V(X) (=V_W(X)+V_B(X))$, $V_W(X)$, and $V_B(X)$ normalized by $Q(X)$ for each layer of each model, based on Fig.~\ref{fig:MV_bar}.
}
\label{fig:MX_VwX_VbX_per_QX}
\end{figure*}

\begin{figure*}[p]
    \centering
    \includegraphics[width=\textwidth]{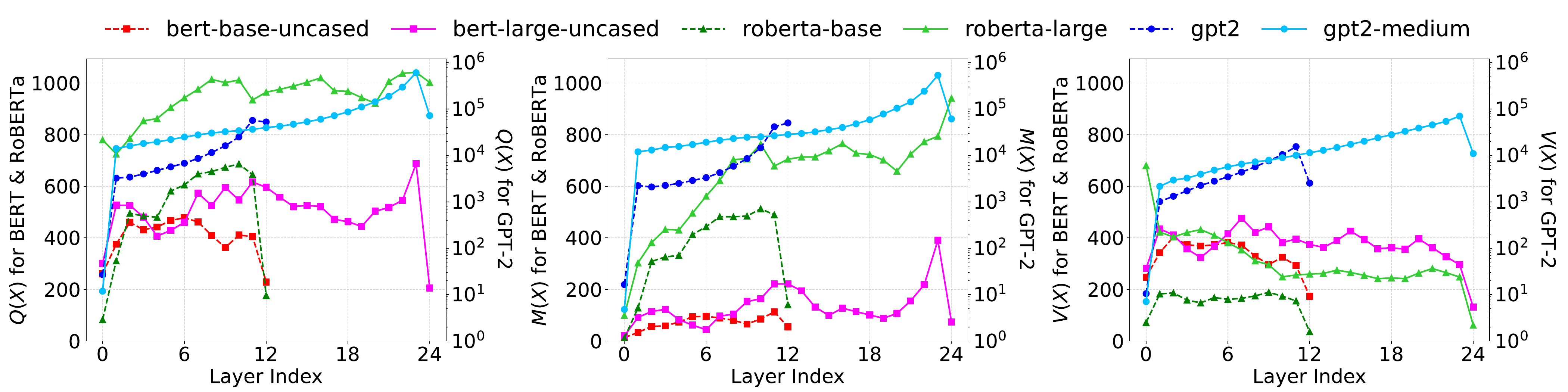}
    \caption{
Values of $Q(X)$, $M(X)$, and $V(X)$ for each layer across the six models. 
For GPT-2, refer to the right vertical axis, as the scale of the values differs from those of BERT and RoBERTa.
}
\label{fig:QX_MX_VX}
\end{figure*}

\begin{figure*}[p]
    \centering
    \includegraphics[width=\textwidth]{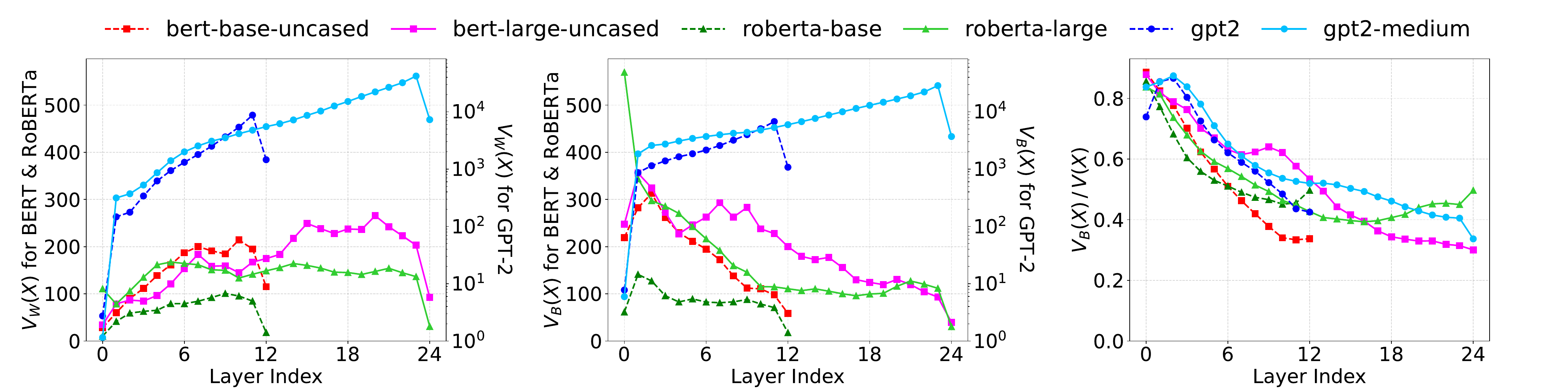}
    \caption{
Values of $V_W(X)$, $V_B(X)$, and $V_B(X)/V(X)$ for each layer across the six models. 
For GPT-2, refer to the right vertical axis for $V_W(X)$ and $V_B(X)$, as the scale of the values differs from those of BERT and RoBERTa. 
Since $V_B(X)/V(X) = 1 - V_W(X)/V(X)$, similar to Fig.~\ref{fig:Vw_per_V}, where $V_W(X)/V(X)$ increases as the layers deepen, $V_B(X)/V(X)$ decreases.
}
\label{fig:Vw_Vb_Vb-per-V}
\end{figure*}

In this section, we present detailed results from Section~\ref{sec:expX}. 
Using the data from the bar graphs in Fig.~\ref{fig:MV_bar}, Fig.~\ref{fig:MX_VwX_VbX_per_QX} shows the normalized values of $M(X)$, $V(X) (=V_W(X)+V_B(X))$, $V_W(X)$, and $V_B(X)$ relative to $Q(X)$ for each layer of each model. 
As observed in Fig.~\ref{fig:MV_bar}, $M(X)/Q(X)$ increases as the layers deepen, while $V(X)/Q(X)$ decreases. 
Furthermore, $V_W(X)/Q(X)$ increases, whereas $V_B(X)/Q(X)$ decreases.
Figure~\ref{fig:QX_MX_VX} shows the values of $Q(X)$, $M(X)$, and $V(X)$ for each layer across the six models.
As seen in Fig.~\ref{fig:MX_VwX_VbX_per_QX}, where $M(X)/Q(X)$ increases as the layers deepen, we observe that $M(X)$ also increases, although the increase varies among the models. 
In contrast, while $V(X)/Q(X)$ decreases in GPT-2 as the layers deepen, the value of $V(X)$ itself increases monotonically, except in the final layer.
It is known that in GPT-2, the norm and standard deviation of the residual stream increase exponentially as the layers deepen~\cite{heimersheim2023residual}, and the results in Fig.~\ref{fig:QX_MX_VX} are consistent with previous work.

Figure~\ref{fig:Vw_Vb_Vb-per-V} shows the values of $V_W(X)$, $V_B(X)$, and $V_B(X)/V(X)$ for each layer across the six models. 
As observed in Fig.~\ref{fig:Vw_per_V}, where $V_W(X)/V(X)$ increases as the layers deepen, we can also see that the value of $V_W(X)$ increases. 
In contrast, in GPT-2, $V_B(X)/V(X)$ decreases as the layers deepen, while the value of $V_B(X)$ itself increases monotonically, except in the final layer.

\section{Explanation of the differences in Transformer layers}\label{app:LN}
\begin{figure}[t!]
    \centering
    \includegraphics[height=8cm,keepaspectratio]{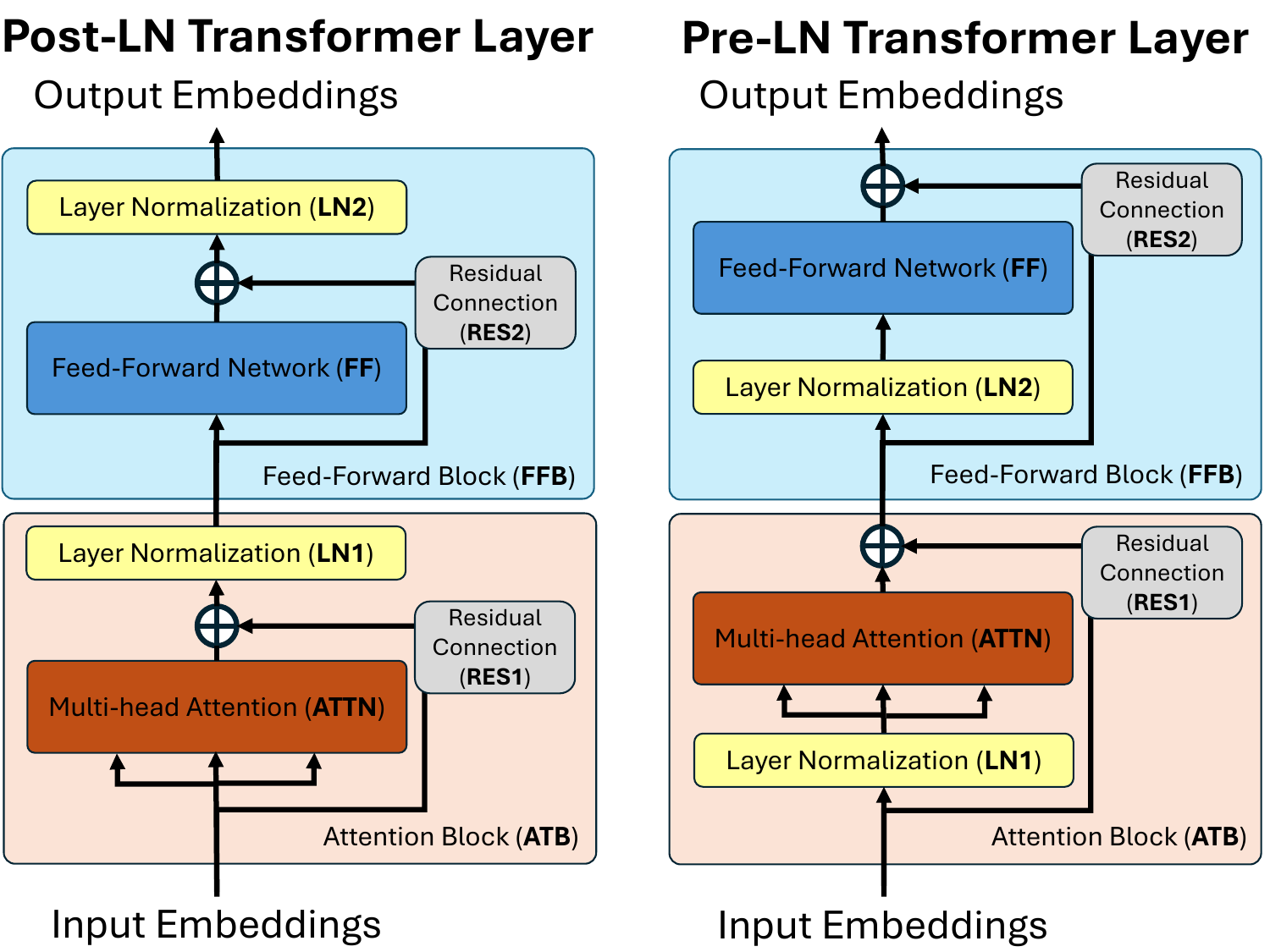}
    \caption{
Figure of the Post-LN and Pre-LN Transformer layer based on \citet{DBLP:conf/iclr/KobayashiKYI24}. 
BERT and RoBERTa have Post-LN layers, while GPT-2 has Pre-LN layers.
}
\label{fig:layer_diff}
\end{figure}

This section introduces the Transformer layers based on the explanation by \citet{DBLP:conf/iclr/KobayashiKYI24}. 
A single Transformer layer~\cite{DBLP:conf/nips/VaswaniSPUJGKP17} consists of four components: multi-head attention (ATTN), feed-forward network (FF), residual connection (RES), and layer normalization (LN), as shown in Fig.~\ref{fig:layer_diff}. 
Following \citet{DBLP:conf/icml/XiongYHZZXZLWL20,DBLP:conf/iclr/KobayashiKYI24}, we classify the layers into Post-LN and Pre-LN Transformer layers based on the position of the LN. 
In the models we used, BERT and RoBERTa have Post-LN layers, while GPT-2 has Pre-LN layers.

A single Transformer layer can be divided into two parts: the Attention Block (ATB), consisting of ATTN, RES1, and LN1, and the Feed-Forward Block (FFB), consisting of FF, RES2, and LN2. 
In this study, we focus on the FFB because we are analyzing the output of each layer. 
For a detailed explanation of the ATB, see \citet{DBLP:conf/iclr/KobayashiKYI24}.
The FF, RES, and LN functions take $\bm{h}\in\mathbb{R}^d$ as input and are defined as follows:
\begin{align}
    FF(\mathbf{h}) &= \bm{W}_2\,\bm{g}(\bm{W}_1\bm{h} + \bm{b}_1) + \bm{b}_2\in\mathbb{R}^{d}, \\
    (RES\circ\bm{v})(\mathbf{h}) &= \bm{v}(\mathbf{h}) + \mathbf{h} \in\mathbb{R}^{d}, \\
    LN(\mathbf{h}) &= \frac{\bm{h}-\text{Mean}(\bm{h})}{\text{Std}(\bm{h})} \odot \bm{\gamma} + \bm{\beta}\in\mathbb{R}^{d}, \label{eq:LNh}
\end{align}
where $\bm{W}_1\in \mathbb{R}^{d'\times d}$ and $\bm{b}_1\in\mathbb{R}^{d'}$ are the weight and bias of the input layer in the FF, $\bm{W}_2\in \mathbb{R}^{d\times d'}$ and $\bm{b}_2\in\mathbb{R}^{d}$ are the weight and bias of the output layer in the FF, and $\bm{\gamma}$ and $\bm{\beta}$ are the weight and bias of the LN.
In addition, $\bm{g}:\mathbb{R}^{d'}\mapsto\mathbb{R}^{d'}$, $\bm{v}:\mathbb{R}^d\mapsto\mathbb{R}^d$, $\text{Mean}: \mathbb{R}^d\mapsto\mathbb{R}$, and $\text{Std}: \mathbb{R}^d\mapsto\mathbb{R}$ represent the activation function, arbitrary vector-valued functions, the mean of the elements, and the standard deviation of the elements, respectively. The operator $\odot$ denotes element-wise multiplication.

We denote the FFB of the Post-LN Transformer layer as $\text{FFB}_\text{Post}$, that of the Pre-LN Transformer layer as $\text{FFB}_\text{Pre}$, and the output of the ATB as $\bm{h}_\text{ATB}\in\mathbb{R}^d$, then we have:
\begin{align}
\text{FFB}_\text{Post}(\bm{h}_\text{ATB})
&= (\text{LN2}\circ\text{RES2}\circ\text{FF})(\bm{h}_\text{ATB}), \label{eq:ffbpost}\\
\text{FFB}_\text{Pre}(\bm{h}_\text{ATB})
&= (\text{RES2}\circ\text{FF}\circ\text{LN2})(\bm{h}_\text{ATB}). \label{eq:ffbpre}
\end{align}

\section{Variation of \texorpdfstring{$Q(X_t)$}{Qt} for the embeddings from the Layer Normalization}
\label{app:theory-layer-nomalization}

Following the definition in (\ref{eq:LNh}), we consider the case that embedding is expressed as
\begin{align*}
  \bm{x} &= LN(\bm{h})\\
         &= \bm{z} \odot \bm{\gamma} + \bm{\beta}\in\mathbb{R}^{d},
\end{align*}
where
\begin{equation}
   \bm{z} := \frac{\bm{h}-\text{Mean}(\bm{h})}{\text{Std}(\bm{h})}. \label{eq:bmz} 
\end{equation}
Corresponding to the sampling $\bm{x}\in X_t$ for the token-wise embedding set, we define the sampling $\bm{z} \in Z_t$, and assume that $Z_t$ is sampled from a distribution $F_t$. Thus
\begin{equation}
    \mathbb{E}_{\bm{z} \in Z_t}\{z_i^k\} = \mathbb{E}_{\bm{z} \sim F_t}\{z_i^k\} + O_p(n_t^{-1/2}), \label{eq:EZt}
\end{equation}
where $n_t = |Z_t|$ is the sample size.
Here we introduce an assumption that the marginal distributions of the elements $z_1,\ldots,z_d$ of $\bm{z} \sim F_t$ are the same; Although this setting does not necessarily reflect reality, we assume it as an ideal scenario.
Since (\ref{eq:bmz}), the sample mean and the sample variance of the elements $z_1,\ldots,z_d$ are zero and one, respectively, we can assume that, for a sufficiently large $d$, the population mean and the population variance of each element $z_i$ in $F_t$ is also zero and one, respectively. Therefore, (\ref{eq:EZt}) with $k=0$ and $k=1$ gives
\[
\mathbb{E}_{\bm{z} \in Z_t}\{z_i\} = O_p(n_t^{-1/2}),\quad
\mathbb{E}_{\bm{z} \in Z_t}\{z_i^2\} = 1 + O_p(n_t^{-1/2}).
\]
Now we look at $Q(X_t)$ for the embeddings from the layer normalization.
\begin{align*}
    Q(X_t) &= \mathbb{E}_{\bm{x}\in X_t}\{\|\bm{x}\|^2\}\\
     &= \mathbb{E}_{\bm{z}\in Z_t}\{\|\bm{z} \odot \bm{\gamma} + \bm{\beta}\|^2\}\\
     &= \mathbb{E}_{\bm{z}\in Z_t} \Bigl\{ 
     \sum_{i=1}^d (\gamma_i z_i + \beta_i)^2
     \Bigr\}\\
     &= \sum_{i=1}^d \Bigl\{
     \gamma_i^2 \mathbb{E}_{\bm{z}\in Z_t} \{z_i^2\}
   + 2\beta_i\gamma_i \mathbb{E}_{\bm{z}\in Z_t} \{z_i\}
   + \beta_i^2  \Bigr\}\\
     &= \sum_{i=1}^d \Bigl\{
     \gamma_i^2 (1 + O_p(n_t^{-1/2}))
   + 2\beta_i\gamma_i O_p(n_t^{-1/2})
   + \beta_i^2  \Bigr\}\\
&= \|\bm{\gamma}\|^2 +  \|\bm{\beta}\|^2 + 
(\|\bm{\gamma}\|^2 + \bm{\beta}^\top\bm{\gamma} )O_p(n_t^{-1/2}).
\end{align*}
This implies that $Q(X_t)\approx \|\bm{\gamma}\|^2 +  \|\bm{\beta}\|^2$ is nearly constant, with variation proportional to $n_t^{-1/2}$.
Since we evaluate the variation of $Q(X_t)$ when sampling $t\in T$, the worst case is $n_0 = \min_{t\in T} n_t$. Therefore, the coefficient of variation (C.V.) is
\[
\textrm{C.V.}(Q(X_t)) = \frac{(\|\bm{\gamma}\|^2 + \bm{\beta}^\top\bm{\gamma} )O(n_0^{-1/2})}{\|\bm{\gamma}\|^2 +  \|\bm{\beta}\|^2 } = O(n_0^{-1/2}).
\]
This C.V. approaches zero as both $d$ and all $n_t$ become larger.

\section{Artifacts of word embeddings represented by the mean of token embeddings}\label{app:wordartifact}
\begin{figure*}[t!]
    \centering
    \includegraphics[width=\textwidth]{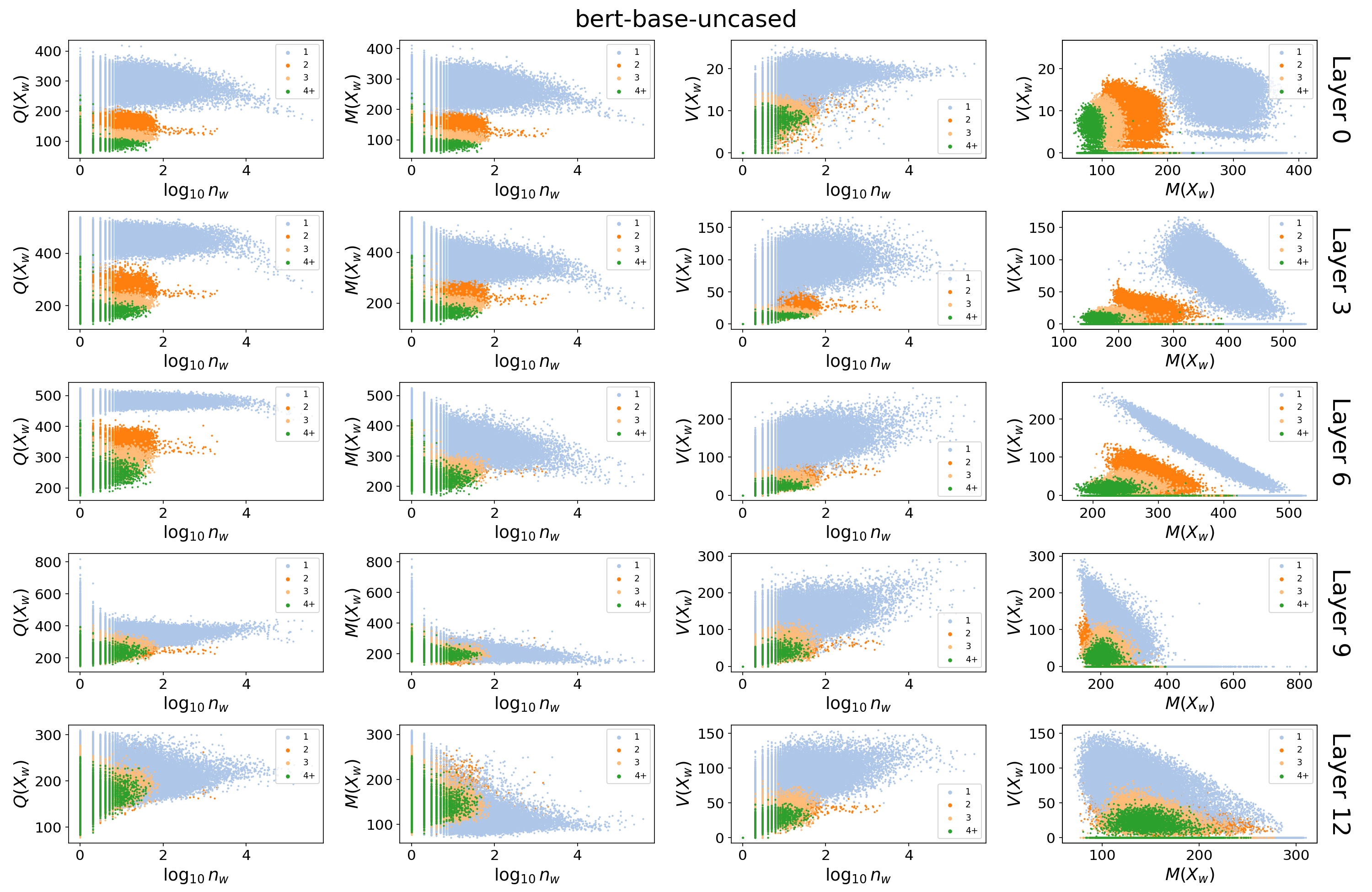}
    \caption{
Results of the same experiments as in Fig.~\ref{fig:QMV_MV_bert_base} using each word embedding set $X_w$, where the embeddings are the means of the token embeddings for \texttt{bert-base-uncased}. 
Words are colored based on the number of tokens produced by BERT tokenization. 
The clustering of words with the same color indicates an artifact caused by taking the means of the token embeddings.
}
\label{fig:word_artifact}
\end{figure*}
In this study, we used token embeddings in our experiments.
In contrast, some previous work investigating the relationship between frequency and embeddings has used {\bf word} embeddings represented by the mean of token embeddings (e.g.,~\citet{DBLP:conf/acl/Wannasuphoprasit23}). 
This section explains the artifacts that can arise when using such word embeddings.

Let $S_w$ be the set of sentences containing the word $w$ in the corpus, and $T_w$ be the set of tokens when word $w$ is tokenized. 
Similar to the token embedding set $X_t$ defined in (\ref{eq:Xt}), we define the embedding set of word $w$ using the $d$-dimensional contextualized embedding model $f$ and $S_w$ as follows:
\begin{align}
    X_w := \left\{\frac{1}{|T_w|}\sum_{t \in T_w}f(s,t) \relmiddle| s\in S_w\right\}.
\end{align}
Using \texttt{bert-base-uncased} as the embedding model, we performed the same experiments on $X_w$ as those in Fig.~\ref{fig:QMV_MV_bert_base}, and we show the results in Fig.~\ref{fig:word_artifact}, with words colored by $|T_w|$. 
Here, unlike Fig.~\ref{fig:QMV_MV_bert_base}, we plotted the actual values of $Q(X_t)$, $M(X_t)$, and $V(X_t)$ instead of using a log scale for better visualization.
As shown in Fig.~\ref{fig:word_artifact}, words with the same number of tokens tend to form clusters, especially in the shallow layers. 
Additionally, the values of $Q(X_w)$, $M(X_w)$, and $V(X_w)$ become smaller as the number of tokens increases. 
This is likely because, as the number of tokens increases, the averaged component values tend to approach zero. 
These results are consistent with previous research~\cite{DBLP:conf/acl/ZhouEJ22}, which showed that the mean norm tends to become smaller as the number of subwords increases. 
With this in mind, we conducted our analysis in this study using token embeddings.

Interestingly, this artifact diminishes as the embeddings become more contextualized in the deeper layers. 
Therefore, in the experiments conducted by~\citet{DBLP:conf/acl/Wannasuphoprasit23}, where only the final layer of \texttt{bert-base-uncased} was used to analyze $\mathbb{E}_{\bm{x}\in X_w} \left\{ \|\bm{x}\|\right\}$, the effect of such artifacts appeared to be relatively small. 
\end{document}